\pdfoutput=1

\documentclass[11pt]{article}

\usepackage[whole]{bxcjkjatype}

\usepackage{booktabs}
\usepackage{tabularx}
\usepackage{multirow}
\usepackage{subcaption}
\usepackage{textcomp}

\usepackage{siunitx}

\usepackage[final]{acl}

\usepackage{times}
\usepackage{latexsym}

\usepackage[T1]{fontenc}

\usepackage[utf8]{inputenc}

\usepackage{microtype}

\usepackage{inconsolata}

\usepackage{graphicx}
\usepackage{tikz}
\usetikzlibrary{positioning,arrows.meta}

\usepackage{pgfplots}
\usepackage{pgfplotstable}
\usetikzlibrary{pgfplots.groupplots}
\pgfplotsset{compat=1.18}

\newcommand{\corC}{blue!10}

\usepackage[most]{tcolorbox}
\newcommand{\xmark}{\textcolor{red}{\ding{55}}} 
\newcolumntype{C}[1]{>{\centering\arraybackslash}p{#1}}
\newcolumntype{L}[1]{>{\raggedright\arraybackslash}p{#1}}

\usepackage{colortbl}
\usepackage{enumitem}
\setlist[itemize]{itemsep=0pt}
\setlist[enumerate]{itemsep=0em}
\usepackage{etoolbox}
\usepackage{pgf}
\usepackage{array}

\definecolor{lightred}{HTML}{CC8685}
\definecolor{darkred}{HTML}{FF0000}

\usepackage{xfp}
\newcommand{\signed}[1]{%
  \begingroup
  \edef\result{\fpeval{#1}}
  \ifdim \result pt>0pt
    $\textcolor{blue}{+\result}$%
  \else
    $\textcolor{red}{\result}$%
  \fi
  \endgroup
}
\newcommand{\colorcell}[1]{#1}

\usepackage{outlines}
\usepackage{comment}
\usepackage{multicol}
\usepackage{textcomp}

\usepackage{outlines}
\usepackage{float}
\usepackage{comment}

\usepackage{listings}
\usepackage{pifont} 

\newcommand{\FourCard}[4]{
\begin{tikzpicture}[scale=0.3]
    \begin{scope}[shift={(0,0)}]
        \fill[gray!30, rounded corners=2pt]
            (0.3,-0.3) rectangle (5.1, 4.8);
        \draw[thick, rounded corners=2pt, fill=white] (0,0) rectangle (4.8,5.0);
        \node at (2.4, 2.5) {\large #1};
    \end{scope}

    \begin{scope}[shift={(5.8,0)}]
        \fill[gray!30, rounded corners=2pt]
            (0.3,-0.3) rectangle (5.1, 4.8);
        \draw[thick, rounded corners=2pt, fill=white] (0,0) rectangle (4.8,5.0);
        \node at (2.4, 2.5) {\large #2};
    \end{scope}

    \begin{scope}[shift={(11.6,0)}]
        \fill[gray!30, rounded corners=2pt]
            (0.3,-0.3) rectangle (5.1, 4.8);
        \draw[thick, rounded corners=2pt, fill=white] (0,0) rectangle (4.8,5.0);
        \node at (2.4, 2.5) {\large #3};
    \end{scope}

    \begin{scope}[shift={(17.4,0)}]
        \fill[gray!30, rounded corners=2pt]
            (0.3,-0.3) rectangle (5.1, 4.8);
        \draw[thick, rounded corners=2pt, fill=white] (0,0) rectangle (4.8,5.0);
        \node at (2.4, 2.5) {\large #4};
    \end{scope}
\end{tikzpicture}
}

\usepackage[most]{tcolorbox}

\newtcolorbox{quoteboxa}[1]{
  title=#1,
  enhanced jigsaw,
  colback=white,
  colbacktitle=white,
  coltitle=black,
  boxrule=0.8pt,
  attach boxed title to top left={xshift=3mm, yshift=-1.5mm},
  box align=top
}

\newcommand{\Success}{\textcolor{green!60!black}{\ding{51}}}
\newcommand{\Fail}{\xmark}

\newcommand{\ErrorExampleBox}[8]{
\begin{tcolorbox}[
    colback=white,      
    colframe=black!50,  
    coltext=black,      
    colbacktitle=gray!15,  
    coltitle=black,        
    title=\parbox{\linewidth}{
        \begin{tabular}{@{}ll}
            \textbf{Type:} \small\texttt{#1} & \textbf{Correct Answer:} \small\texttt{#2}
        \end{tabular}
    }
]
\begin{tabular}{@{}>{\raggedleft\arraybackslash}p{4ex}L{0.85\linewidth}@{}}
\textbf{Rule:} & #3 \\
\textbf{1.} & #4 \\
\textbf{2.} & #5 \\
\textbf{3.} & #6 \\
\textbf{4.} & #7 \\
\end{tabular}
#8
\end{tcolorbox}
}

\title{Evaluation of Deontic Conditional Reasoning in Large Language Models: The Case of Wason's Selection Task}

\author{Hirohiko Abe\textsuperscript{1}, Kentaro Ozeki\textsuperscript{1,2}, Risako Ando\textsuperscript{1}, Takanobu Morishita\textsuperscript{1},\\ \textbf{Koji Mineshima\textsuperscript{1}, Mitsuhiro Okada\textsuperscript{1}}\\
  \textsuperscript{1}Keio University, Tokyo, Japan\\ 
  \textsuperscript{2}University of Tokyo, Tokyo, Japan\\ 
  \texttt{\{hirohiko-abe,risakochaan,morishita\}@keio.jp}\quad
  \texttt{kentaro.ozeki@gmail.com}\\
  \texttt{\{minesima,okada\}@abelard.flet.keio.ac.jp}}

\begin{document}
\maketitle

\begin{abstract}
As large language models (LLMs) advance in linguistic competence, their reasoning abilities are gaining increasing attention.
In humans, reasoning often performs well in domain specific settings, particularly in normative rather than purely formal contexts.
Although prior studies have compared LLM and human reasoning, the domain specificity of LLM reasoning remains underexplored.
In this study, we introduce a new Wason Selection Task dataset that explicitly encodes deontic modality to systematically distinguish deontic from descriptive conditionals, and use it to examine LLMs' conditional reasoning under deontic rules.
We further analyze whether observed error patterns are better explained by confirmation bias (a tendency to seek rule-supporting evidence) or by matching bias (a tendency to ignore negation and select items that lexically match elements of the rule).
Results show that, like humans, LLMs reason better with deontic rules and display matching-bias-like errors.
Together, these findings suggest that the performance of LLMs varies systematically across rule types and that their error patterns can parallel well-known human biases in this paradigm.

\end{abstract}

\section{Introduction}
\label{sec:introduction}

As large language models (LLMs) advance in linguistic capabilities, their reasoning abilities are gaining increasing attention.
Among various forms of reasoning, inference based on conditional rules of the form ``if $p$ then $q$,'' known as \textbf{conditional reasoning}, is especially important.
Such reasoning underlies domains such as moral and legal judgment and has significant real-world implications.

In cognitive science, empirical studies have shown that humans tend to be more proficient in conditional reasoning within specific domains, especially those involving norms, compared to purely formal reasoning~\cite{griggs1982elusive,cheng1985pragmatic,manktelow1991social}.
This phenomenon, known as \textbf{domain specificity}, suggests that human reasoning relies on different cognitive mechanisms across domains~\cite{davies1995logical,fiddick2004domains}.
Such characteristics have been explained as outcomes of evolutionary adaptation~\cite{cosmides1989logic,cosmides-Tooby1992}.
Whether LLMs trained with large-scale neural networks exhibit domain specificity in their reasoning abilities remains an open question.

Research in cognitive science has also pointed out that human errors in conditional reasoning stem from \textit{matching bias}~\cite{evans1973matching,evans1998matching}, a tendency to focus on the words appearing in the rule itself, rather than from confirmation bias~\cite{Wason1966-WASR}, a tendency to \textit{confirm} a given rule.
Although prior research compared human and LLM reasoning based on insights from cognitive science~\cite{seals-shalin-2024-evaluating,dasgupta2022language,ozeki2024exploring}, it remains unclear which bias best accounts for LLM errors. To our knowledge, confirmation bias and matching bias have not been systematically compared within a unified experimental design.

In this paper, we evaluate LLMs' conditional reasoning using the Wason Selection Task~\cite{Wason1966-WASR}, a standard paradigm in cognitive psychology.
We use a new dataset that explicitly encodes deontic modality, enabling systematic comparisons between deontic and descriptive rules.\footnote{The dataset is publicly available at
\url{https://github.com/kmineshima/NeuBAROCO}.}
LLMs achieve higher accuracy on deontic rules, paralleling patterns of human performance.
Moreover, on deontic rules, their error patterns are more consistent with matching bias than with confirmation bias.
These results suggest domain-specificity effects in LLM conditional reasoning and error patterns that align with matching bias.

Our contributions are summarized as follows:
\begin{enumerate}\setlength{\itemsep}{-0.2em}
    \item We introduce a new Wason Selection Task dataset with explicit deontic encoding.
    \item We provide a comprehensive and up-to-date evaluation of LLMs using this dataset.
    \item We clarify parallels in human and LLM performance across deontic versus descriptive rules.
    \item We compare confirmation and matching bias as explanations of LLM errors, finding stronger support for matching bias.
\end{enumerate}

\section{Selection Task and Domain Specificity}
\label{sec:selection}

The Wason Selection Task is a widely studied paradigm in cognitive science for probing the gap between formal logic and human intuitive reasoning about conditionals.
It assesses whether a rule expressed in the form ``if $p$ then $q$'' holds true.
Participants are presented with four cards that represent the affirmation or negation of the rule's antecedent and consequent, and asked to select which cards must be turned over to verify the rule.

In the example below (Rule 1), the logically correct answer is to select the cards ``7'' and ``d,'' which have the potential to falsify the rule.
However, experimental studies in cognitive psychology have widely observed that fewer than 10\% of participants choose this correct combination~\cite[Chap.~4]{evans1993human}.

\begin{figure}[h]
\textbf{Rule 1 (Descriptive)}: If a card has an odd number on one side, then the opposite side has an uppercase letter.
\smallskip

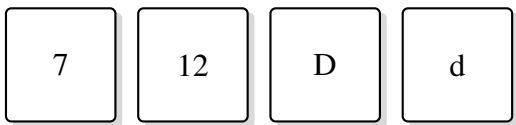
\centering
\FourCard{7}{12}{D}{d}
\end{figure}

In our study, we refer to rules whose antecedent and consequent are arbitrarily paired and lack a meaningful commonsense connection (as in the original Wason task) as \textit{descriptive}, in contrast to the \textit{deontic} rules discussed below.

It is well established that participants perform better on tasks with concrete content, especially when the content involves normative concepts such as obligations or permissions, rather than abstract or general materials~\cite{cheng1985pragmatic,manktelow1991social}.
For instance, in Rule 2, participants are more likely to select the correct cards (``blood'' and ``no gloves'') than in Rule 1.

\begin{figure}[h]
\textbf{Rule 2 (Deontic)}: If blood is spilled, the nurse must wear gloves.
\smallskip

\centering
\FourCard{\small{Blood}}{\small{No blood}}{\small{Gloves}}{\small{No gloves}}
\end{figure}

Although the precise mechanisms behind this phenomenon remain debated, a leading hypothesis suggests that human reasoning operates through domain-specificity mechanisms that are specialized according to the type of content involved~\cite{cosmides1989logic,cosmides-Tooby1992}.

Recent NLP research has also focused on conditional reasoning using the Wason Selection Task~\cite{dasgupta2022language, seals-shalin-2024-evaluating}.
While these studies investigate LLM performance on the Wason Selection Task, they do not explicitly frame their analyses in terms of {domain specificity} in the cognitive-science sense we adopt.

In \citet{dasgupta2022language}, the authors primarily investigate content effects (belief biases) in LLMs.
They do not address the structural distinction between reasoning about deontic versus descriptive rules, which is central to our definition of domain specificity.

\citet{seals-shalin-2024-evaluating} compare LLM performance on the Wason Selection Task with \textit{social} and \textit{non-social} rules.
They report that the evaluated models tend to perform better when reasoning about social rules, though this advantage is smaller than would be expected in humans.
However, they do not provide an explicit criterion for systematically distinguishing social from non-social categories.%

We extend this line of research by focusing on the \textbf{modality} expressed in the rules.
Specifically, to control the type of rule and enable a more systematic evaluation, we classify rules into two categories: \textit{descriptive} and \textit{deontic}.

\textit{Deontic} rules express norms that regulate actions, such as obligations and prohibitions.
For example, the rules containing obligation markers (e.g., ``must'' and ``need to'') or prohibition markers (e.g., ``must not'' and ``not allowed'') are classified as {deontic}.
In contrast, \textit{descriptive} rules (e.g., Rule 1) state generalizations without reference to actions or deontic modal expressions.
This modality-based distinction helps define a clearer boundary between deontic and descriptive rules and supports more consistent classification.

\begin{table*}[h!]
    \centering
    \scalebox{0.9}{
    \begin{tabular}{cclcccc}
        \toprule
        \textbf{Rule Polarity} & \textbf{Pattern} & \textbf{Example Rule} & \textbf{TA} & \textbf{FA} & \textbf{TC} & \textbf{FC}\\
        \midrule
        \texttt{Pos-Pos} &
        $p \to q$ &
        If the number is prime, then the letter is lowercase. &
        \fcolorbox{black}{\corC}{\texttt{7}} & \texttt{8} & \fcolorbox{black}{white}{\texttt{b}} & \colorbox{\corC}{\texttt{B}}
        \\
        \texttt{Pos-Neg} &
        $p \to \neg q$ &
        If the number is prime, then the letter is \textbf{not} lowercase. &
        \fcolorbox{black}{\corC}{\texttt{7}} & \texttt{8} & \texttt{B} & \fcolorbox{black}{\corC}{\texttt{b}}
        \\
        \texttt{Neg-Pos} &
        $\neg p \to q$ &
        If the number is \textbf{not} prime, then the letter is lowercase. &
        \colorbox{\corC}{\texttt{8}} & \fcolorbox{black}{white}{\texttt{7}} & \fcolorbox{black}{white}{\texttt{b}} & \colorbox{\corC}{\texttt{B}}
        \\
        \texttt{Neg-Neg} &
        $\neg p \to \neg q$ &
        If the number is \textbf{not} prime, then the letter is \textbf{not} lowercase. &
        \colorbox{\corC}{\texttt{8}} & \fcolorbox{black}{white}{\texttt{7}} & \texttt{B} & \fcolorbox{black}{\corC}{\texttt{b}}
        \\
        \bottomrule
    \end{tabular}
    }
    \caption{%
    List of patterns involving negation.
    \texttt{Pos}: affirmative, \texttt{Neg}: negated.
    \textbf{TA}: case where the antecedent is true (True Antecedent),
    \textbf{FA}: case where the antecedent is false (False Antecedent),
    \textbf{TC}: case where the consequent is true (True Consequent),
    \textbf{FC}: case where the consequent is false (False Consequent).
    Cells highlighted in \colorbox{\corC}{blue} indicate logically correct selections.
    \raisebox{0.3ex}{\tiny\fbox{\rule{0pt}{1.5ex}\rule{1ex}{0pt}}} indicates options predicted by matching bias.
    Confirmation bias predicts selections of both \textbf{TA} and \textbf{TC}.
    }
    \label{tab:wason-schema}
\end{table*}

\section{Confirmation Bias and Matching Bias}
\label{sec:confirmation}

A typical response to the task based on Rule 1 is to select the cards ``7'' and ``D.''
Schematically, for a rule in the form ``if $p$, then $q$,'' the logically correct answer is to choose cases where $p$ is true and $q$ is false; that is, cases that would \emph{falsify} the rule.
However, people commonly select cases where both $p$ and $q$ are true, i.e., cases that \emph{confirm} the rule.
Wason's original hypothesis was that this tendency stems from a confirmation bias~\cite{Wason1966-WASR}.
On the other hand, later studies have suggested that the error pattern observed in human responses to the Wason Selection Task may be better explained by \textit{matching bias}, rather than \textit{confirmation bias}~\cite{evans1973matching,evans1998matching}.
This can be tested by using the same card arrangement as in Rule 1, with the rule replaced by its \textit{negated} form.

\begin{figure}[h]
\noindent
\textbf{Rule 3 (Descriptive-negation)}: If a card has an odd number on one side, then the opposite side is \textbf{not} an uppercase letter.

\medskip

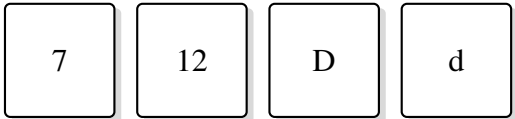
\centering
\FourCard{7}{12}{D}{d}
\end{figure}

\noindent
In this case, the ``D'' card represents not a confirming instance, but a falsifying one.
Nevertheless, participants have still been observed to select the ``7'' and ``D'' cards in such tasks~\cite{evans1998matching}.
This suggests that people tend to select cards that match the elements appearing in the rule regardless of whether those elements are negated, rather than choosing based on the truth of the antecedent and consequent (as in confirmation bias.)
This tendency indicates \textit{matching bias}.

To investigate whether LLMs replicate confirmation bias or matching bias, we designed four types of \emph{rule polarity}, based on whether the antecedent and/or consequent includes negation (see Table~\ref{tab:wason-schema}).
If a model exhibits confirmation bias, it should consistently select the cards where the antecedent is true (\textbf{TA}) and the consequent is true (\textbf{TC}), regardless of whether negation appears in the rule.
If a model exhibits matching bias, it should prefer options that directly match the symbols $p$ and $q$ in the rule, rather than those corresponding to $\neg p$ or $\neg q$.

Matching bias can be characterized as a tendency to overlook negation and select cards that lexically match the terms mentioned in the rule.
Accordingly, testing for matching bias amounts to testing sensitivity to negation in conditional reasoning.
Such analyses therefore probe whether LLMs can correctly handle negation, a long-standing challenge for Transformer-based models~\cite{ettinger2020}.
Incorporating negation into the Wason task allows for a more precise identification of the sources of reasoning biases.

\section{Experiments}
\label{sec:experiments}

\subsection{Dataset}
\label{ssec:dataset}

To evaluate LLM performance on the Wason Selection Task across different modalities and polarities, we constructed an evaluation dataset in which each problem consists of a rule, a set of four numbered cards corresponding to the rule, the numbers of the two correct choices, and a gold label.
The dataset contains a total of 160 problems, including 80 problems with \textit{deontic} rules and 80 problems with \textit{descriptive} rules.
Each of the four rule polarity patterns (\texttt{Pos-Pos}, \texttt{Pos-Neg}, \texttt{Neg-Pos}, \texttt{Neg-Neg}) consists of 20 problems.
See Appendix~\ref{appendix:dataset} for details of the dataset construction.

\subsection{Models and Settings}
\label{ssec:models-settings}

In the experiment, we evaluate five families of open-weight models, including both reasoning and non-reasoning models.
For reasoning models, we evaluate gpt-oss 20B/120B \cite{openai2025gpt-oss} and Qwen 3 14B/32B \cite{yang2025qwen3}.
For non-reasoning models, we evaluate Gemma 3 4B/12B/27B \cite{ms2025gemma3}, Llama 3.3 70B \cite{grattafiori2024llama3}, and OLMo 2 32B \cite{olmo2024olmo2}.
For all non-reasoning models, instruction-tuned variants are used.
Compared with \citet{seals-shalin-2024-evaluating}, which evaluated earlier, relatively small non-reasoning models and reported low performance compared to humans, we evaluate more recent models, including models explicitly trained or tuned for reasoning.

The basic (Zero-Shot) prompt used in the experiment is provided in Appendix~\ref{appendix:prompts}.
For the Few-Shot prompt, we include one exemplar question and answer for a deontic rule and one for a descriptive rule. The answers in these exemplars are intentionally different from the correct answers to the corresponding experimental questions.
The CoT prompt adopts the Zero-Shot Chain-of-Thought method~\cite{kojima2022large}.
For non-reasoning models, the initial maximum output token limit is set to 10 for the Zero- and Few-Shot prompts, and to 6,000 for the CoT prompt.
For reasoning models, the initial maximum output token limit is set to 6,000. When this limit is insufficient, it is incrementally increased by 1,000 (up to a maximum of 11,000 in total) and the request is retried.
The temperature is set to 0 (more deterministic outputs) for non-reasoning models and to the recommended values for reasoning models.
All other hyperparameters are kept at their default values.

We use accuracy as the evaluation metric, computed using an exact-match criterion for this multi-select task.
A response is counted as correct if and only if the model selects all and only the correct options.

\subsection{Results and Discussion}
\label{ssec:results}

This section presents results on domain specificity and reasoning bias.
Examples of observed reasoning errors are provided in Appendix~\ref{appendix:errors}.

\begin{table}[th]
\centering

\begin{subtable}{\linewidth}
\setlength\tabcolsep{3pt}
\centering
\resizebox{\columnwidth}{!}{
\begin{tabular}{lccc}
\toprule
\textbf{Model} &
\textbf{Zero} & \textbf{Few} & \textbf{CoT} \\
\midrule
\multicolumn{4}{l}{\textbf{Reasoning Models}} \\
\midrule
gpt-oss-20b
    & \colorcell{67.5} (\signed{\fpeval{67.5-40.0}})
    & \colorcell{91.2} (\signed{\fpeval{91.2-50.0}})
    & n/a \\
gpt-oss-120b
    & \colorcell{100.0} (\signed{\fpeval{100.0-70.0}})
    & \colorcell{98.8} (\signed{\fpeval{98.8-68.8}})
    & n/a \\
qwen3-14b
    & \colorcell{97.5} (\signed{\fpeval{97.5-52.5}})
    & \colorcell{98.8} (\signed{\fpeval{98.8-61.3}})
    & n/a \\
qwen3-32b
    & \colorcell{98.8} (\signed{\fpeval{98.8-60.0}})
    & \colorcell{100.0} (\signed{\fpeval{100.0-75.0}})
    & n/a \\
\midrule
\multicolumn{4}{l}{\textbf{Non-Reasoning Models}} \\
\midrule
gemma3-4b 
    & \colorcell{17.5} (\signed{\fpeval{17.5-12.5}})
    & \colorcell{21.2} (\signed{\fpeval{21.2-8.7}})
    & \colorcell{38.8} (\signed{\fpeval{38.8-18.8}}) \\
gemma3-12b 
    & \colorcell{58.8} (\signed{\fpeval{58.8-28.7}})
    & \colorcell{78.7} (\signed{\fpeval{78.7-37.5}})
    & \colorcell{43.8} (\signed{\fpeval{43.8-10.0}}) \\
gemma3-27b 
    & \colorcell{77.5} (\signed{\fpeval{77.5-38.8}})
    & \colorcell{76.2} (\signed{\fpeval{76.2-47.5}})
    & \colorcell{62.5} (\signed{\fpeval{62.5-37.5}}) \\
llama-3.3-70b 
    & \colorcell{62.5} (\signed{\fpeval{62.5-52.5}})
    & \colorcell{78.7} (\signed{\fpeval{78.7-62.5}})
    & \colorcell{83.8} (\signed{\fpeval{83.8-45.0}}) \\
olmo-2-0325-32b 
    & \colorcell{36.2} (\signed{\fpeval{36.2-21.2}})
    & \colorcell{38.8} (\signed{\fpeval{38.8-22.5}})
    & \colorcell{57.5} (\signed{\fpeval{57.5-41.2}}) \\
    \bottomrule
\end{tabular}
}
\end{subtable}

\caption{%
Accuracy (\%) by prompt type for deontic rule problems. Values in parentheses show the difference in accuracy from the descriptive rule condition (positive if deontic > descriptive). \textbf{Zero} = Zero-Shot, \textbf{Few} = Few-Shot, \textbf{CoT} = Chain-of-Thought.
}
\label{tab:results-overall}
\end{table}

\begin{table}[t]
    \centering
    \resizebox{\columnwidth}{!}{%
    \begin{tabular}{c c c c c c}
    \toprule
    \textbf{Polarity} & \textbf{Pattern} & \textbf{TA} & \textbf{FA} & \textbf{TC} & \textbf{FC} \\
    \midrule
    \texttt{pos-pos} & $p \to q$ & \fcolorbox{black}{\corC}{100.0} & 0.0 & \fcolorbox{black}{white}{45.0} & \colorbox{\corC}{55.0} \\
    \texttt{pos-neg} & $p \to \lnot q$ & \fcolorbox{black}{\corC}{100.0} & 0.0 & 5.0 & \fcolorbox{black}{\corC}{95.0} \\
    \texttt{neg-pos} & $\lnot p \to q$ & \colorbox{\corC}{80.0} & \fcolorbox{black}{white}{20.0} & \fcolorbox{black}{white}{60.0} & \colorbox{\corC}{40.0} \\
    \texttt{neg-neg} & $\lnot p \to \lnot q$ & \colorbox{\corC}{90.0} & \fcolorbox{black}{white}{10.0} & 10.0 & \fcolorbox{black}{\corC}{90.0} \\
    \bottomrule
    \end{tabular}}
    \caption{Selection percentages (\%) for each option in the deontic rule problems (gpt-oss-20b, Zero-Shot setting).
    As with Table~\ref{tab:wason-schema}, cells highlighted in \colorbox{\corC}{blue} indicate logically correct selections.
    \raisebox{0.3ex}{\tiny\fbox{\rule{0pt}{1.5ex}\rule{1ex}{0pt}}} indicates options predicted by matching bias.
    Confirmation bias predicts selections of both \textbf{TA} and \textbf{TC}.%
    }
    \label{tab:results-gpt-oss-20b}
\end{table}

\begin{table}[th]
\centering

\begin{subtable}{\linewidth}
\setlength\tabcolsep{2pt}
\centering
\resizebox{\columnwidth}{!}{
    \begin{tabular}{lcccccccc}
      \toprule
      \multirow{2}{*}{\textbf{Model}} & \multicolumn{2}{c}{\textbf{TA}} & \multicolumn{2}{c}{\textbf{FA}} & \multicolumn{2}{c}{\textbf{TC}} & \multicolumn{2}{c}{\textbf{FC}} \\
      & \textbf{$p$} & \textbf{$\neg p$} & \textbf{$p$} & \textbf{$\neg p$} & \textbf{$q$} & \textbf{$\neg q$} & \textbf{$q$} & \textbf{$\neg q$} \\
      \midrule
\multicolumn{9}{l}{\textbf{Reasoning Models}} \\
\midrule
gpt-oss-20b & \textbf{100.0} & 85.0 & \textbf{15.0} & 0.0 & \textbf{52.5} & 7.5 & \textbf{92.5} & 47.5 \\
gpt-oss-120b & {100.0} & {100.0} & {0.0} & {0.0} & {0.0} & {0.0} & {100.0} & {100.0} \\
qwen3-14b & \textbf{100.0} & 95.0 & \textbf{5.0} & 0.0 & {0.0} & {0.0} & {100.0} & {100.0} \\
qwen3-32b & \textbf{100.0} & 97.5 & {0.0} & {0.0} & {0.0} & {0.0} & {100.0} & {100.0} \\
\midrule
\multicolumn{9}{l}{\textbf{Non-Reasoning Models}} \\
\midrule
gemma-3-4b & \textbf{100.0} & 95.0 & 2.5 & \textbf{17.5} & \textbf{15.0} & 2.5 & \textbf{32.5} & 2.5 \\
gemma-3-12b & 77.5 & \textbf{95.0} & 10.0 & \textbf{55.0} & \textbf{15.0} & 2.5 & \textbf{85.0} & 57.5 \\
gemma-3-27b & \textbf{100.0} & 85.0 & \textbf{25.0} & 5.0 & \textbf{12.5} & 7.5 & \textbf{92.5} & 75.0 \\
llama-3.3-70b & \textbf{100.0} & 85.0 & \textbf{20.0} & 0.0 & \textbf{62.5} & 0.0 & \textbf{100.0} & 47.5 \\
olmo-2-0325-32b & \textbf{70.0} & 65.0 & 0.0 & 0.0 & \textbf{85.0} & 2.5 & \textbf{100.0} & 25.0 \\
\bottomrule
      \end{tabular}
}
\caption{Deontic Rule}
\end{subtable}

\bigskip

\begin{subtable}{\linewidth}
\setlength\tabcolsep{2pt}
\centering
\resizebox{\columnwidth}{!}{
    \begin{tabular}{lcccccccc}
      \toprule
      \multirow{2}{*}{\textbf{Model}} & \multicolumn{2}{c}{\textbf{TA}} & \multicolumn{2}{c}{\textbf{FA}} & \multicolumn{2}{c}{\textbf{TC}} & \multicolumn{2}{c}{\textbf{FC}} \\
      & \textbf{$p$} & \textbf{$\neg p$} & \textbf{$p$} & \textbf{$\neg p$} & \textbf{$q$} & \textbf{$\neg q$} & \textbf{$q$} & \textbf{$\neg q$} \\
      \midrule
\multicolumn{9}{l}{\textbf{Reasoning Models}} \\
\midrule
gpt-oss-20b & \textbf{100.0} & 92.5 & 5.0 & \textbf{15.0} & 25.0 & \textbf{30.0} & 70.0 & \textbf{75.0} \\
gpt-oss-120b & 100.0 & 100.0 & 2.5 & \textbf{20.0} & 5.0 & \textbf{25.0} & 95.0 & 95.0 \\
qwen3-14b & \textbf{100.0} & 97.5 & 0.0 & \textbf{7.5} & 27.5 & \textbf{30.0} & \textbf{87.5} & 77.5 \\
qwen3-32b & \textbf{100.0} & 100.0 & 0.0 & \textbf{12.5} & 17.5 & \textbf{32.5} & 90.0 & 90.0 \\
\midrule
\multicolumn{9}{l}{\textbf{Non-Reasoning Models}} \\
\midrule
gemma-3-4b & \textbf{100.0} & 47.5 & \textbf{87.5} & 0.0 & \textbf{47.5} & 0.0 & \textbf{37.5} & 5.0 \\
gemma-3-12b & \textbf{97.5} & 82.5 & \textbf{37.5} & 32.5 & \textbf{30.0} & 20.0 & \textbf{55.0} & 20.0 \\
gemma-3-27b & \textbf{100.0} & 90.0 & \textbf{45.0} & 7.5 & \textbf{70.0} & 2.5 & \textbf{80.0} & 5.0 \\
llama-3.3-70b & \textbf{100.0} & 92.5 & 7.5 & \textbf{37.5} & \textbf{50.0} & 15.0 & \textbf{77.5} & 45.0 \\
olmo-2-0325-32b & \textbf{87.5} & 27.5 & \textbf{50.0} & 0.0 & \textbf{67.5} & 20.0 & \textbf{80.0} & 32.5 \\
      \bottomrule
      \end{tabular}
}
\caption{Descriptive Rule}
\end{subtable}

\caption{
Percentage (\%) of types of options selected by the models, relative to rule polarity (Zero-Shot setting).
For the meanings of \textbf{TA, FA, TC, FC}, refer to Section~\ref{sec:confirmation} and Table~\ref{tab:wason-schema}.
\textbf{Boldface} indicates the higher value between $p$ and $\neg p$, or $q$ and $\neg q$, in each respective column (when applicable).
}
\label{tab:results-bias}

\end{table}

\subsubsection{Domain Specificity}

Table~\ref{tab:results-overall} shows the differences in performance between the deontic and descriptive rule problems.
Across all models and prompt types, accuracy is higher on deontic rule problems than on descriptive rule problems, with improvements ranging from 5.0 to 41.2\%.
The degree of the difference between the two types of rule tasks varied across models.
Overall, these findings suggest that LLMs exhibit domain-specificity effects in reasoning with deontic rules, though the extent of this effect appears to depend on factors including model size.

\subsubsection{Confirmation Bias versus Matching Bias}

Table~\ref{tab:results-gpt-oss-20b} shows a model's selection percentages for each option under each rule polarity,
following the format of Table~\ref{tab:wason-schema}.
For example, when the rule has the polarity of \texttt{Pos-Pos} ($p \to q$), the model selects the \textbf{TA} (True Antecedent) option in 100.0\% of cases (top-left cell of Table~\ref{tab:results-gpt-oss-20b}).
Within the same row (i.e., under the same rule pattern), the selection percentages for \textbf{FA}, \textbf{TC}, and \textbf{FC} are 0.0\%, 45.0\%, and 55.0\%, respectively.
Note that each option is selected independently of the others.
Corresponding tables for all models are provided in the Appendix~\ref{appendix:results}.

As stated in Section~\ref{sec:confirmation}, if confirmation bias were to be replicated, we would expect a high selection rate for the options corresponding to \textbf{TA} and \textbf{TC}, regardless of whether they are in affirmative or negative form.
However, as shown in Table~\ref{tab:results-gpt-oss-20b},
while \textbf{TA} exceeds \textbf{FA}, \textbf{TC} does not exceed \textbf{FC}. Therefore, these results do not provide evidence for confirmation bias.

To assess whether models exhibit matching bias, Table~\ref{tab:results-bias} reorganizes the data following the analytic framework presented in \citet{evans1998matching} for human experiments.
Here, the matching bias is evaluated by comparing \textbf{relative preferences} among \textit{p} vs.\ \textit{¬p} and \textit{q} vs.\ \textit{¬q} \textbf{within each option}.
For example, in the \textbf{TC} column in Table~\ref{tab:results-gpt-oss-20b}, identifying whether the model prefers \textit{q} (as reflected in the \texttt{Pos-Pos} and \texttt{Neg-Pos} rows) over \textit{¬q} (the \texttt{Pos-Neg} and \texttt{Neg-Neg} rows) is crucial for diagnosing matching bias.
Table~\ref{tab:results-bias} highlights these contrasts concisely; accordingly, the larger of the two percentages is boldfaced, visually marking the preferred choice within each pair.
Note that the labels \textit{p}, \textit{¬p}, \textit{q}, and \textit{¬q} in the column headers indicate the polarity of the \textbf{selected option type} (i.e., which card was chosen), rather than the antecedent and consequent forms of the rules (e.g., $p \to q$ and $\neg p \to \neg q$).

As discussed in Section~\ref{sec:confirmation},
matching bias predicts a preference for $p$ over $\neg p$ and $q$ over $\neg q$, irrespective of whether the antecedent or consequent in the options is true or false.
As shown in Table~\ref{tab:results-bias}, our results are broadly consistent with this prediction.

One exception arises for reasoning models in the descriptive condition, where $\neg p$ is consistently preferred to $p$ in \textbf{FA} and $\neg q$ to $q$ in \textbf{TC}.
Since only \textbf{TA} and \textbf{FC} are required to check for violations, selecting \textbf{FA} or \textbf{TC} yields an incorrect response under our exact-match scoring, and these options are chosen only infrequently compared with \textbf{TA} and \textbf{FC}.

Finally, note that matching bias and logical correctness are not mutually exclusive.
In particular, selecting \textit{q} in \textbf{FC}, as predicted by matching bias, is logically valid.
Thus, a matching-bias-consistent preference in \textbf{FC} does not entail irrational reasoning but can even increase accuracy.

\section{Conclusion}
\label{sec:conclusion}
This study systematically evaluated LLMs' conditional reasoning using the Wason Selection Task.
As shown in Section~\ref{ssec:results}, the results suggest that LLMs exhibit domain-specificity effects and that their error tendencies are more consistent with matching bias than confirmation bias.

From the perspective of explaining this behavior in LLMs, further investigation into the specific domains of specialization and the mechanisms behind them remains an open question.
In addition, mechanistic analysis to determine the causal origins of those behaviors is left for future work.
Although this study centered on the Wason Selection Task, future work should extend the analysis to other forms of conditional reasoning and distinct types of reasoning beyond conditional reasoning, within the modality-based framework proposed in this study.
Additionally, a systematic evaluation of LLMs' reasoning about permission and obligation constitutes an important avenue for further investigation.

\section*{Limitations}

While our study aims to provide a comprehensive analysis of deontic conditional reasoning in LLMs, several limitations should be noted.

First, this study examined the Wason Selection Task as a case study of conditional reasoning. However, whether these findings extend to conditional reasoning more broadly remains an open question for future work.

In addition, while our study focused on deontic modality to examine domain specificity, further research could also consider other ways of classifying domains.
Relatedly, a systematic evaluation of LLMs' reasoning about permission and obligation remains an important direction for further investigation.

More broadly, while our study focused on characterizing the observable reasoning behavior of LLMs in a controlled variant of the Wason Selection Task, mechanistic analyses that determine the causal origins of these behaviors, such as architectural inductive biases, properties of the training data, or specific exposure to Wason type problems, are valuable topics for future work.

Finally, because LLMs are continuously evolving, our findings may not generalize to future models with improved reasoning mechanisms. Model performance can change with updates to training data, architecture, and fine tuning strategies.
Moreover, differences in fine tuning, prompt engineering, and instruction following abilities may contribute to performance variation, which makes it challenging to isolate the effects of model architecture alone.

\section*{Acknowledgment}
We thank the anonymous reviewers for their insightful comments and suggestions, which have improved the paper.
This work is partially supported by JST CREST Grant Number JPMJCR2114, Keio University Global Research Institute (KGRI) Challenge Grant, and JSPS Kakenhi Grant Numbers JP24K00004, JP21K00016, JP21H00467, JP23K20416, and JP21K18339.

\bibliography{custom}

@article{evans1973matching,
  title={Matching bias in the selection task},
  author={Evans, J St BT and Lynch, J St},
  journal={British Journal of Psychology},
  volume={64},
  number={3},
  pages={391--397},
  year={1973},
  publisher={Wiley Online Library}
}

@article{olmo2024olmo2,
  title={2 {OLMo} 2 Furious},
  author={{Team OLMo}},
  journal={arXiv preprint arXiv:2501.00656},
  year={2024}
}

@article{grattafiori2024llama3,
  title={The llama 3 herd of models},
  author={Grattafiori, Aaron and Dubey, Abhimanyu and Jauhri, Abhinav and Pandey, Abhinav and Kadian, Abhishek and Al-Dahle, Ahmad and Letman, Aiesha and Mathur, Akhil and Schelten, Alan and Vaughan, Alex and others},
  journal={arXiv preprint arXiv:2407.21783},
  year={2024}
}

@article{ms2025gemma3,
  title={Gemma 3 technical report},
  author={{Gemma Team}},
  journal={arXiv preprint arXiv:2503.19786},
  year={2025}
}

@article{yang2025qwen3,
  title={Qwen3 technical report},
  author={Yang, An and Li, Anfeng and Yang, Baosong and Zhang, Beichen and Hui, Binyuan and Zheng, Bo and Yu, Bowen and Gao, Chang and Huang, Chengen and Lv, Chenxu and others},
  journal={arXiv preprint arXiv:2505.09388},
  year={2025}
}

@article{openai2025gpt-oss,
  title={gpt-oss-120b \& gpt-oss-20b model card},
  author={OpenAI},
  journal={arXiv preprint arXiv:2508.10925},
  year={2025}
}

@article{ettinger2020,
    author = {Ettinger, Allyson},
    title = {What {BERT} Is Not: Lessons from a New Suite of Psycholinguistic Diagnostics for Language Models},
    journal = {Transactions of the Association for Computational Linguistics},
    volume = {8},
    pages = {34-48},
    year = {2020},
    month = {01},
    issn = {2307-387X},
    doi = {10.1162/tacl_a_00298},
    url = {https://doi.org/10.1162/tacl\_a\_00298},
    eprint = {https://direct.mit.edu/tacl/article-pdf/doi/10.1162/tacl\_a\_00298/1923116/tacl\_a\_00298.pdf},
}

@article{evans1998matching,
  title={Matching bias in conditional reasoning: Do we understand it after 25 years?},
  author={Evans, Jonathan St BT},
  journal={Thinking \& Reasoning},
  volume={4},
  number={1},
  pages={45--110},
  year={1998},
  publisher={Taylor \& Francis}
}

@article{manktelow1991social,
  title={Social roles and utilities in reasoning with deontic conditionals},
  author={Manktelow, Ken I and Over, David E},
  journal={Cognition},
  volume={39},
  number={2},
  pages={85--105},
  year={1991},
  publisher={Elsevier}
}

@article{davies1995logical,
  title={Logical reasoning and domain specificity: A critique of the social exchange theory of reasoning},
  author={Davies, Paul Sheldon and Fetzer, James H and Foster, Thomas R},
  journal={Biology and Philosophy},
  volume={10},
  pages={1--37},
  year={1995},
  publisher={Springer}
}

@article{griggs1982elusive,
  title={The elusive thematic-materials effect in {W}ason's selection task},
  author={Griggs, Richard A and Cox, James R},
  journal={British Journal of Psychology},
  volume={73},
  number={3},
  pages={407--420},
  year={1982},
  publisher={Wiley Online Library}
}

@article{cheng1985pragmatic,
  title={Pragmatic reasoning schemas},
  author={Cheng, Patricia W and Holyoak, Keith J},
  journal={Cognitive psychology},
  volume={17},
  number={4},
  pages={391--416},
  year={1985},
  publisher={Elsevier}
}

@article{kojima2022large,
  title={Large language models are zero-shot reasoners},
  author={Kojima, Takeshi and Gu, Shixiang Shane and Reid, Machel and Matsuo, Yutaka and Iwasawa, Yusuke},
  journal={Advances in neural information processing systems},
  volume={35},
  pages={22199--22213},
  year={2022}
}

@inproceedings{seals-shalin-2024-evaluating,
    title = "Evaluating the Deductive Competence of Large Language Models",
    author = "Spencer M. Seals and Valerie L. Shalin",
    editor = "Duh, Kevin  and
      Gomez, Helena  and
      Bethard, Steven",
    booktitle = "Proceedings of the 2024 Conference of the North American Chapter of the Association for Computational Linguistics: Human Language Technologies (Volume 1: Long Papers)",
    month = jun,
    year = "2024",
    address = "Mexico City, Mexico",
    publisher = "Association for Computational Linguistics",
    url = "https://aclanthology.org/2024.naacl-long.476/",
    doi = "10.18653/v1/2024.naacl-long.476",
    pages = "8614--8630"
}

@article{cosmides1989logic,
  title={The logic of social exchange: Has natural selection shaped how humans reason? Studies with the Wason selection task},
  author={Cosmides, Leda},
  journal={Cognition},
  volume={31},
  number={3},
  pages={187--276},
  year={1989},
  publisher={Elsevier}
}

@inproceedings{ozeki2024exploring,
title={Exploring Reasoning Biases in Large Language Models Through Syllogism: Insights from the {N}eu{BAROCO} Dataset},
author={Ozeki, Kentaro and Ando, Risako  and
      Morishita, Takanobu  and
      Abe, Hirohiko  and
      Mineshima, Koji  and
      Okada, Mitsuhiro},
booktitle={Findings of the Association for Computational Linguistics: ACL 2024},
year={2024}
}

@article{dasgupta2022language,
    author = {Lampinen, Andrew K and Dasgupta, Ishita and Chan, Stephanie C Y and Sheahan, Hannah R and Creswell, Antonia and Kumaran, Dharshan and McClelland, James L and Hill, Felix},
    title = {Language models, like humans, show content effects on reasoning tasks},
    journal = {PNAS Nexus},
    volume = {3},
    number = {7},
    pages = {page233},
    year = {2024},
    issn = {2752-6542},
    doi = {10.1093/pnasnexus/pgae233},
    url = {https://doi.org/10.1093/pnasnexus/pgae233},
    eprint = {https://academic.oup.com/pnasnexus/article-pdf/3/7/pgae233/58651606/pgae233.pdf},
}

@book{evans1993human,
  title={Human Reasoning: The Psychology of Deduction},
  author={Evans, Jonathan St.B. T.  and Newstead, Stephen E. and Byrne, Ruth M. J.},
  year={1993},
  publisher={Psychology Press}
}

@incollection{Wason1966-WASR,
	author = {Peter C. Wason},
	booktitle = {New Horizons in Psychology},
	editor = {Peter C. Wason},
	pages = {135--151},
	publisher = {Penguin Books},
	title = {Reasoning},
	year = {1966}
}

@incollection{cosmides-Tooby1992,
    author = {Cosmides, Leda and Tooby, John},
    title = {Cognitive Adaptations for Social Exchange},
    booktitle = {The Adapted Mind: Evolutionary Psychology and the Generation of Culture},
    publisher = {Oxford University Press},
    year = {1992}
}

@article{fiddick2004domains,
  title={Domains of deontic reasoning: Resolving the discrepancy between the cognitive and moral reasoning literatures},
  author={Fiddick, Laurence},
  journal={The Quarterly Journal of Experimental Psychology Section A},
  volume={57},
  number={3},
  pages={447--474},
  year={2004},
  publisher={Taylor \& Francis}
}

\appendix

\section{Dataset Details}
\label{appendix:dataset}

\begin{table*}[t]
    \centering
    \scalebox{0.9}{
    \begin{tabularx}{\textwidth}{ccXX}
        \toprule
        \textbf{Rule Polarity} & \textbf{Pattern} &
        \textbf{Descriptive} & \textbf{Deontic}\\
        \midrule
        \texttt{Pos-Pos} &
        $p \to q$ &
        If a number is prime, then the opposite side has a lowercase letter. & If the lamp is on, then the visitor must enter the room.
        \\
        \texttt{Pos-Neg} &
        $p \to \neg q$ &
        If a number is prime, then the opposite side \textbf{does not} have a lowercase letter. & If the lamp is on, then the visitor \textbf{must not} enter the room.
        \\
        \texttt{Neg-Pos} &
        $\neg p \to q$ &
        If a number is \textbf{not} prime, then the opposite side has a lowercase letter. & If the lamp is \textbf{not} on, then the visitor must enter the room.
        \\
        \texttt{Neg-Neg} &
        $\neg p \to \neg q$ &
        If a number is \textbf{not} prime, then the opposite side \textbf{does not} have a lowercase letter. & If the lamp is \textbf{not} on, then the visitor \textbf{must not} enter the room.
        \\
        \bottomrule
    \end{tabularx}
    }
    \caption{Example rules of descriptive and deontic cases.}
    \label{tab:negation}
\end{table*}

We created templates as described in Section~\ref{sec:selection}.
Then, we used Gemini 1.5 Pro to instantiate these templates. 
We manually checked and substantially adjusted the cases to ensure quality.

We created 20 \textit{deontic} and 20 \textit{descriptive} cases with a \texttt{Pos–Pos} polarity pattern.
By manually adding negation to the antecedent and/or consequent, we generated the cases in other polarity patterns, namely \texttt{Pos-Neg}, \texttt{Neg-Pos}, \texttt{Neg-Neg}.
Formally, the application of negation can be summarized as shown in Table~\ref{tab:negation}.

Regarding the application of negation to the consequent in \textit{deontic} cases, obligations were treated as affirmative whereas prohibitions (negations of permission) as negative.
This is due to the fact that
the rules expressing permission or unnecessity (negation of obligation) were excluded from the dataset because the cards to be selected differ in these modal types from obligation and prohibition (negation of permission).
For example, when the rule is one of permission, such as Rule 4,
there is no card that should be selected.
This is because, even if the antecedent holds, there is no requirement to perform the action described in the consequent, and conversely, if the antecedent does not hold, it is not logically entailed that the action in the consequent must not be performed.

\bigskip

\noindent
\textbf{Rule 4 (Permission)}: If the lamp is on, then the visitor is allowed to enter the room.
\bigskip

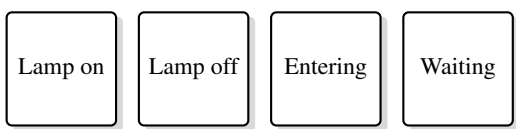
\begin{center}
\FourCard{\small{Lamp on}}{\small{Lamp off}}{\small{Entering}}{\small{Waiting}}
\end{center}

\bigskip

\noindent
The same applies to the unnecessity (negation of obligation), as in ``If the lamp is on, then the visitor needs not enter the room.''
In this case as well, there is no card that should be selected.
In the cases of permission and unnecessity (negation of obligation), unlike in the cases of obligation and prohibition (negation of permission), the cards satisfying the antecedent and those negating the consequent are not the correct choices.
To avoid inconsistency with respect to the correct answers, these types were excluded from the study.

\section{Prompts}
\label{appendix:prompts}

An example of a basic (Zero-Shot) prompt is as follows:

\begin{quoteboxa}{Basic Prompt}
To verify if the following rule holds true, which of the four cards below should you flip over?
Please do not turn over more than required.
Answer with zero or more of the numbers 1 to 4, separated by commas, and output nothing else.\\

Rule: If a symbol is a spiral, then the opposite side is a primary color.\\
Card 1: Spiral\\
Card 2: Triangle\\
Card 3: Red\\
Card 4: Green\\

Answer:1,4

\end{quoteboxa}

\section{Error Examples}
\label{appendix:errors}

Examples of errors for the descriptive and deontic rule problems, including three polarity combinations (\texttt{Pos–Pos}, \texttt{Neg–Pos}, and \texttt{Neg–Neg}), are presented in Figures~\ref{fig:errors-deontic-1}, \ref{fig:errors-deontic-2}, \ref{fig:errors-epistemic-1}, and \ref{fig:errors-epistemic-2}.

Figure~\ref{fig:errors-deontic-1} shows an example of a \textit{deontic} rule with \texttt{Pos-Pos} polarity.
Several models (e.g., {gpt-oss-20b}) incorrectly selected option~3, which corresponds to the consequent $q$, instead of option~4 (\textit{Guard leaving}), which represents $\lnot q$ and thus the actual violation when $p$ holds.
This error pattern is consistent with both confirmation bias and matching bias.

Figure~\ref{fig:errors-deontic-2} shows an example of a \textit{deontic} rule with \texttt{Neg-Pos} polarity.
Some models (e.g., gpt-oss-20b) again incorrectly selected option~3, which corresponds to the consequent $q$, instead of option~4 ($\lnot q$, \textit{Document not filed}), which represents the actual violation when $p$ holds.
This error pattern is compatible with both confirmation bias and matching bias.
On the other hand, {gemma-3-27b} and {llama-3.3-70b} incorrectly selected option~1 (\textit{Car in front of the building}) instead of option~2 (\textit{Car in the parking lot}).
This error reflects a tendency to select options that lexically match the antecedent while ignoring negation, and is therefore more consistent with matching bias than with confirmation bias.

Figure~\ref{fig:errors-epistemic-1} presents a \textit{descriptive} rule with \texttt{Pos-Pos} polarity.
Some models (e.g., gemma-3-27b) incorrectly picked option~3 (\textit{Violin}), which matches the consequent $q$, even though the correct choice is to check option~4 (\textit{Bicycle}, i.e., $\lnot q$) to verify whether the rule is violated when the antecedent $p$ holds.
This error pattern, which also appears in Figure~\ref{fig:errors-deontic-1} for the \textit{deontic} condition, is compatible with both confirmation bias and matching bias.

Figure~\ref{fig:errors-epistemic-2} shows a \textit{descriptive} rule with \texttt{Neg-Neg} polarity.
The logically correct response is to select option~2 (\%), which satisfies the antecedent, and option~3 (\textit{Sad}), which negates the consequent.
Some models, including gemma-3-4b and gemma-3-12b, selected options~1 (\textmusicalnote) and option~3 (\textit{Sad}).
Selecting option~1 instead of option~2 suggests that the model ignored negation in the antecedent and instead chose the lexically matching option.
This behavior is therefore consistent with matching bias rather than confirmation bias.
Note that selecting the negation of the consequent is required, so choosing option~3 ($\neg q$) is both logically correct and compatible with matching bias.
Some models (gpt-oss-20b and gpt-oss-120b) additionally included option~4 ($q$) alongside option~3, yielding a mixed response pattern that is not straightforward to interpret as confirmation bias.

\begin{figure}[H]
\centering
\ErrorExampleBox{Deontic}{1, 4}
{If it is sunny, then the guard is obliged to unlock the door.}
{Sunny}
{Rainy}
{Guard unlocking the door}
{Guard leaving}
{
\begin{center}
\begin{tabular}{@{}l C{2cm}@{}}
\toprule
\textbf{Model} & \textbf{Prediction} \\
\midrule
gpt-oss-20b   & \Fail \ 1, 3 \\
gpt-oss-120b  & \Success \ 1, 4 \\
qwen3-14b    & \Success \ 1, 4 \\
qwen3-32b    & \Success \ 1, 4 \\
gemma-3-4b    & \Fail \ 1 \\
gemma-3-12b   & \Fail \ 1, 2 \\
gemma-3-27b   & \Fail \ 1, 3 \\
llama-3.3-70b & \Fail \ 1, 3 \\
olmo-2-0325-32b  & \Fail \ 3 \\
\bottomrule
\end{tabular}
\end{center}
}
\caption{An example of error for a deontic rule with \texttt{Pos-Pos} polarity.}
\label{fig:errors-deontic-1}
\end{figure}

\begin{figure}[H]
\centering
\ErrorExampleBox{Deontic}{2, 4}
{If the car is not in front of the building, then the lawyer needs to file the document with the court.}
{Car in front of the building}
{Car in the parking lot}
{Lawyer filing}
{Document not filed}
{
\begin{center}
\begin{tabular}{@{}l C{2cm}@{}}
\toprule
\textbf{Model} & \textbf{Prediction} \\
\midrule
gpt-oss-20b   & \Fail \ 2, 3 \\
gpt-oss-120b  & \Success \ 2, 4 \\
qwen3-14b    & \Success \ 2, 4 \\
qwen3-32b    & \Success \ 2, 4 \\
gemma-3-4b    & \Success \ 2, 4 \\
gemma-3-12b   & \Success \ 2, 4 \\
gemma-3-27b   & \Fail \ 1, 4 \\
llama-3.3-70b & \Fail \ 1, 3, 4 \\
olmo-2-0325-32b  & \Fail \ 3, 4 \\
\bottomrule
\end{tabular}
\end{center}
}
\caption{An example of error for a deontic rule with \texttt{Neg-Pos} polarity.}
\label{fig:errors-deontic-2}
\end{figure}

\begin{figure}[H]
\centering
\ErrorExampleBox{Descriptive}{1, 4}
{If a number is a multiple of 3, then the opposite side shows a musical instrument.}
{9}
{11}
{Violin}
{Bicycle}
{
\begin{center}
\begin{tabular}{@{}l C{2cm}@{}}
\toprule
\textbf{Model} & \textbf{Prediction} \\
\midrule
gpt-oss-20b   & \Success \ 1, 4 \\
gpt-oss-120b  & \Success \ 1, 4 \\
qwen3-14b    & \Success \ 1, 4 \\
qwen3-32b    & \Success \ 1, 4 \\
gemma-3-4b    & \Fail \ 1 \\
gemma-3-12b   & \Fail \ 2, 4 \\
gemma-3-27b   & \Fail \ 1, 3 \\
llama-3.3-70b & \Fail \ 1, 2, 4 \\
olmo-2-0325-32b  & \Fail \ 1, 3 \\
\bottomrule
\end{tabular}
\end{center}
}
\caption{An example of error for a descriptive rule with \texttt{Pos-Pos} polarity.}
\label{fig:errors-epistemic-1}
\end{figure}

\begin{figure}[H]
\centering
\ErrorExampleBox{Descriptive}{2, 3}
{If a card does not show a musical note, then the opposite side does not contain a word describing a feeling.}
{\textmusicalnote}
{\%}
{Sad}
{Fast}
{
\begin{center}
\begin{tabular}{@{}l C{2cm}@{}}
\toprule
\textbf{Model} & \textbf{Prediction} \\
\midrule
gpt-oss-20b   & \Fail \ 2, 3, 4 \\
gpt-oss-120b  & \Fail \ 2, 3, 4 \\
qwen3-14b    & \Success \ 2, 3 \\
qwen3-32b    & \Success \ 2, 3 \\
gemma-3-4b    & \Fail \ 1, 3 \\
gemma-3-12b   & \Fail \ 1, 3 \\
gemma-3-27b   & \Fail \ 1, 3 \\
llama-3.3-70b & \Success \ 2, 3 \\
olmo-2-0325-32b  & \Fail \ 3, 4 \\
\bottomrule
\end{tabular}
\end{center}
}
\caption{An example of error for a descriptive rule with \texttt{Neg-Neg} polarity.}
\label{fig:errors-epistemic-2}
\end{figure}

\clearpage

\onecolumn

\section{Supplemental Results}
\label{appendix:results}

Tables~\ref{tab:results-gpt-oss-20b-base} to \ref{tab:results-olmo-2-0325-32b-cot} present the selection percentages (\%) by each model under different prompting conditions (Zero-Shot, Few-Shot, CoT), for all rule polarity patterns in both deontic and descriptive rule tasks.
See Table~\ref{tab:results-gpt-oss-20b} for the description of the table format.
\begin{table}[H]
\centering
\begin{subtable}[h]{0.48\textwidth}
\centering
\resizebox{\linewidth}{!}{%
\begin{tabular}{c c c c c c}
    \toprule
    \textbf{Polarity} & \textbf{Pattern} & \textbf{TA} & \textbf{FA} & \textbf{TC} & \textbf{FC} \\
    \midrule
    \texttt{pos-pos} & $p \to q$ & \fcolorbox{black}{\corC}{100.0} & 0.0 & \fcolorbox{black}{white}{45.0} & \colorbox{\corC}{55.0} \\
    \texttt{pos-neg} & $p \to \lnot q$ & \fcolorbox{black}{\corC}{100.0} & 0.0 & 5.0 & \fcolorbox{black}{\corC}{95.0} \\
    \texttt{neg-pos} & $\lnot p \to q$ & \colorbox{\corC}{80.0} & \fcolorbox{black}{white}{20.0} & \fcolorbox{black}{white}{60.0} & \colorbox{\corC}{40.0} \\
    \texttt{neg-neg} & $\lnot p \to \lnot q$ & \colorbox{\corC}{90.0} & \fcolorbox{black}{white}{10.0} & 10.0 & \fcolorbox{black}{\corC}{90.0} \\
    \bottomrule
\end{tabular}
}
\caption{Deontic}
\end{subtable}
\hfill
\begin{subtable}[h]{0.48\textwidth}
\centering
\resizebox{\linewidth}{!}{%
\begin{tabular}{c c c c c c}
    \toprule
    \textbf{Polarity} & \textbf{Pattern} & \textbf{TA} & \textbf{FA} & \textbf{TC} & \textbf{FC} \\
    \midrule
    \texttt{pos-pos} & $p \to q$ & \fcolorbox{black}{\corC}{100.0} & 25.0 & \fcolorbox{black}{white}{10.0} & \colorbox{\corC}{70.0} \\
    \texttt{pos-neg} & $p \to \lnot q$ & \fcolorbox{black}{\corC}{100.0} & 5.0 & 0.0 & \fcolorbox{black}{\corC}{55.0} \\
    \texttt{neg-pos} & $\lnot p \to q$ & \colorbox{\corC}{95.0} & \fcolorbox{black}{white}{10.0} & \fcolorbox{black}{white}{40.0} & \colorbox{\corC}{80.0} \\
    \texttt{neg-neg} & $\lnot p \to \lnot q$ & \colorbox{\corC}{90.0} & \fcolorbox{black}{white}{0.0} & 60.0 & \fcolorbox{black}{\corC}{85.0} \\
    \bottomrule
\end{tabular}
}
\caption{Descriptive}
\end{subtable}
\caption{gpt-oss-20b (Zero-Shot)}
\label{tab:results-gpt-oss-20b-base}
\end{table}

\begin{table}[H]
\centering
\begin{subtable}[h]{0.48\textwidth}
\centering
\resizebox{\linewidth}{!}{%
\begin{tabular}{c c c c c c}
    \toprule
    \textbf{Polarity} & \textbf{Pattern} & \textbf{TA} & \textbf{FA} & \textbf{TC} & \textbf{FC} \\
    \midrule
    \texttt{pos-pos} & $p \to q$ & \fcolorbox{black}{\corC}{100.0} & 0.0 & \fcolorbox{black}{white}{0.0} & \colorbox{\corC}{100.0} \\
    \texttt{pos-neg} & $p \to \lnot q$ & \fcolorbox{black}{\corC}{100.0} & 0.0 & 15.0 & \fcolorbox{black}{\corC}{100.0} \\
    \texttt{neg-pos} & $\lnot p \to q$ & \colorbox{\corC}{100.0} & \fcolorbox{black}{white}{0.0} & \fcolorbox{black}{white}{5.0} & \colorbox{\corC}{100.0} \\
    \texttt{neg-neg} & $\lnot p \to \lnot q$ & \colorbox{\corC}{100.0} & \fcolorbox{black}{white}{0.0} & 15.0 & \fcolorbox{black}{\corC}{100.0} \\
    \bottomrule
\end{tabular}
}
\caption{Deontic}
\end{subtable}
\hfill
\begin{subtable}[h]{0.48\textwidth}
\centering
\resizebox{\linewidth}{!}{%
\begin{tabular}{c c c c c c}
    \toprule
    \textbf{Polarity} & \textbf{Pattern} & \textbf{TA} & \textbf{FA} & \textbf{TC} & \textbf{FC} \\
    \midrule
    \texttt{pos-pos} & $p \to q$ & \fcolorbox{black}{\corC}{100.0} & 35.0 & \fcolorbox{black}{white}{5.0} & \colorbox{\corC}{85.0} \\
    \texttt{pos-neg} & $p \to \lnot q$ & \fcolorbox{black}{\corC}{100.0} & 5.0 & 10.0 & \fcolorbox{black}{\corC}{85.0} \\
    \texttt{neg-pos} & $\lnot p \to q$ & \colorbox{\corC}{90.0} & \fcolorbox{black}{white}{15.0} & \fcolorbox{black}{white}{35.0} & \colorbox{\corC}{95.0} \\
    \texttt{neg-neg} & $\lnot p \to \lnot q$ & \colorbox{\corC}{90.0} & \fcolorbox{black}{white}{0.0} & 60.0 & \fcolorbox{black}{\corC}{95.0} \\
    \bottomrule
\end{tabular}
}
\caption{Descriptive}
\end{subtable}
\caption{gpt-oss-20b (Few-Shot)}
\label{tab:results-gpt-oss-20b-kshot}
\end{table}

\begin{table}[H]
\centering
\begin{subtable}[h]{0.48\textwidth}
\centering
\resizebox{\linewidth}{!}{%
\begin{tabular}{c c c c c c}
    \toprule
    \textbf{Polarity} & \textbf{Pattern} & \textbf{TA} & \textbf{FA} & \textbf{TC} & \textbf{FC} \\
    \midrule
    \texttt{pos-pos} & $p \to q$ & \fcolorbox{black}{\corC}{100.0} & 0.0 & \fcolorbox{black}{white}{0.0} & \colorbox{\corC}{100.0} \\
    \texttt{pos-neg} & $p \to \lnot q$ & \fcolorbox{black}{\corC}{100.0} & 0.0 & 0.0 & \fcolorbox{black}{\corC}{100.0} \\
    \texttt{neg-pos} & $\lnot p \to q$ & \colorbox{\corC}{100.0} & \fcolorbox{black}{white}{0.0} & \fcolorbox{black}{white}{0.0} & \colorbox{\corC}{100.0} \\
    \texttt{neg-neg} & $\lnot p \to \lnot q$ & \colorbox{\corC}{100.0} & \fcolorbox{black}{white}{0.0} & 0.0 & \fcolorbox{black}{\corC}{100.0} \\
    \bottomrule
\end{tabular}
}
\caption{Deontic}
\end{subtable}
\hfill
\begin{subtable}[h]{0.48\textwidth}
\centering
\resizebox{\linewidth}{!}{%
\begin{tabular}{c c c c c c}
    \toprule
    \textbf{Polarity} & \textbf{Pattern} & \textbf{TA} & \textbf{FA} & \textbf{TC} & \textbf{FC} \\
    \midrule
    \texttt{pos-pos} & $p \to q$ & \fcolorbox{black}{\corC}{100.0} & 40.0 & \fcolorbox{black}{white}{0.0} & \colorbox{\corC}{95.0} \\
    \texttt{pos-neg} & $p \to \lnot q$ & \fcolorbox{black}{\corC}{100.0} & 0.0 & 5.0 & \fcolorbox{black}{\corC}{100.0} \\
    \texttt{neg-pos} & $\lnot p \to q$ & \colorbox{\corC}{100.0} & \fcolorbox{black}{white}{5.0} & \fcolorbox{black}{white}{10.0} & \colorbox{\corC}{95.0} \\
    \texttt{neg-neg} & $\lnot p \to \lnot q$ & \colorbox{\corC}{100.0} & \fcolorbox{black}{white}{0.0} & 45.0 & \fcolorbox{black}{\corC}{90.0} \\
    \bottomrule
\end{tabular}
}
\caption{Descriptive}
\end{subtable}
\caption{gpt-oss-120b (Zero-Shot)}
\label{tab:results-gpt-oss-120b-base}
\end{table}

\begin{table}[H]
\centering
\begin{subtable}[h]{0.48\textwidth}
\centering
\resizebox{\linewidth}{!}{%
\begin{tabular}{c c c c c c}
    \toprule
    \textbf{Polarity} & \textbf{Pattern} & \textbf{TA} & \textbf{FA} & \textbf{TC} & \textbf{FC} \\
    \midrule
    \texttt{pos-pos} & $p \to q$ & \fcolorbox{black}{\corC}{100.0} & 0.0 & \fcolorbox{black}{white}{0.0} & \colorbox{\corC}{100.0} \\
    \texttt{pos-neg} & $p \to \lnot q$ & \fcolorbox{black}{\corC}{100.0} & 0.0 & 0.0 & \fcolorbox{black}{\corC}{100.0} \\
    \texttt{neg-pos} & $\lnot p \to q$ & \colorbox{\corC}{100.0} & \fcolorbox{black}{white}{0.0} & \fcolorbox{black}{white}{0.0} & \colorbox{\corC}{95.0} \\
    \texttt{neg-neg} & $\lnot p \to \lnot q$ & \colorbox{\corC}{100.0} & \fcolorbox{black}{white}{0.0} & 0.0 & \fcolorbox{black}{\corC}{100.0} \\
    \bottomrule
\end{tabular}
}
\caption{Deontic}
\end{subtable}
\hfill
\begin{subtable}[h]{0.48\textwidth}
\centering
\resizebox{\linewidth}{!}{%
\begin{tabular}{c c c c c c}
    \toprule
    \textbf{Polarity} & \textbf{Pattern} & \textbf{TA} & \textbf{FA} & \textbf{TC} & \textbf{FC} \\
    \midrule
    \texttt{pos-pos} & $p \to q$ & \fcolorbox{black}{\corC}{100.0} & 45.0 & \fcolorbox{black}{white}{0.0} & \colorbox{\corC}{95.0} \\
    \texttt{pos-neg} & $p \to \lnot q$ & \fcolorbox{black}{\corC}{100.0} & 0.0 & 5.0 & \fcolorbox{black}{\corC}{95.0} \\
    \texttt{neg-pos} & $\lnot p \to q$ & \colorbox{\corC}{100.0} & \fcolorbox{black}{white}{15.0} & \fcolorbox{black}{white}{15.0} & \colorbox{\corC}{90.0} \\
    \texttt{neg-neg} & $\lnot p \to \lnot q$ & \colorbox{\corC}{100.0} & \fcolorbox{black}{white}{0.0} & 35.0 & \fcolorbox{black}{\corC}{100.0} \\
    \bottomrule
\end{tabular}
}
\caption{Descriptive}
\end{subtable}
\caption{gpt-oss-120b (Few-Shot)}
\label{tab:results-gpt-oss-120b-kshot}
\end{table}

\begin{table}[H]
\centering
\begin{subtable}[h]{0.48\textwidth}
\centering
\resizebox{\linewidth}{!}{%
\begin{tabular}{c c c c c c}
    \toprule
    \textbf{Polarity} & \textbf{Pattern} & \textbf{TA} & \textbf{FA} & \textbf{TC} & \textbf{FC} \\
    \midrule
    \texttt{pos-pos} & $p \to q$ & \fcolorbox{black}{\corC}{100.0} & 0.0 & \fcolorbox{black}{white}{0.0} & \colorbox{\corC}{100.0} \\
    \texttt{pos-neg} & $p \to \lnot q$ & \fcolorbox{black}{\corC}{100.0} & 0.0 & 0.0 & \fcolorbox{black}{\corC}{100.0} \\
    \texttt{neg-pos} & $\lnot p \to q$ & \colorbox{\corC}{95.0} & \fcolorbox{black}{white}{5.0} & \fcolorbox{black}{white}{0.0} & \colorbox{\corC}{100.0} \\
    \texttt{neg-neg} & $\lnot p \to \lnot q$ & \colorbox{\corC}{95.0} & \fcolorbox{black}{white}{5.0} & 0.0 & \fcolorbox{black}{\corC}{100.0} \\
    \bottomrule
\end{tabular}
}
\caption{Deontic}
\end{subtable}
\hfill
\begin{subtable}[h]{0.48\textwidth}
\centering
\resizebox{\linewidth}{!}{%
\begin{tabular}{c c c c c c}
    \toprule
    \textbf{Polarity} & \textbf{Pattern} & \textbf{TA} & \textbf{FA} & \textbf{TC} & \textbf{FC} \\
    \midrule
    \texttt{pos-pos} & $p \to q$ & \fcolorbox{black}{\corC}{100.0} & 10.0 & \fcolorbox{black}{white}{5.0} & \colorbox{\corC}{65.0} \\
    \texttt{pos-neg} & $p \to \lnot q$ & \fcolorbox{black}{\corC}{100.0} & 5.0 & 10.0 & \fcolorbox{black}{\corC}{85.0} \\
    \texttt{neg-pos} & $\lnot p \to q$ & \colorbox{\corC}{95.0} & \fcolorbox{black}{white}{0.0} & \fcolorbox{black}{white}{50.0} & \colorbox{\corC}{90.0} \\
    \texttt{neg-neg} & $\lnot p \to \lnot q$ & \colorbox{\corC}{100.0} & \fcolorbox{black}{white}{0.0} & 50.0 & \fcolorbox{black}{\corC}{90.0} \\
    \bottomrule
\end{tabular}
}
\caption{Descriptive}
\end{subtable}
\caption{qwen3-14b (Zero-Shot)}
\label{tab:results-qwen3-14b-base}
\end{table}

\begin{table}[H]
\centering
\begin{subtable}[h]{0.48\textwidth}
\centering
\resizebox{\linewidth}{!}{%
\begin{tabular}{c c c c c c}
    \toprule
    \textbf{Polarity} & \textbf{Pattern} & \textbf{TA} & \textbf{FA} & \textbf{TC} & \textbf{FC} \\
    \midrule
    \texttt{pos-pos} & $p \to q$ & \fcolorbox{black}{\corC}{100.0} & 0.0 & \fcolorbox{black}{white}{0.0} & \colorbox{\corC}{100.0} \\
    \texttt{pos-neg} & $p \to \lnot q$ & \fcolorbox{black}{\corC}{100.0} & 0.0 & 0.0 & \fcolorbox{black}{\corC}{100.0} \\
    \texttt{neg-pos} & $\lnot p \to q$ & \colorbox{\corC}{95.0} & \fcolorbox{black}{white}{0.0} & \fcolorbox{black}{white}{0.0} & \colorbox{\corC}{100.0} \\
    \texttt{neg-neg} & $\lnot p \to \lnot q$ & \colorbox{\corC}{100.0} & \fcolorbox{black}{white}{0.0} & 0.0 & \fcolorbox{black}{\corC}{100.0} \\
    \bottomrule
\end{tabular}
}
\caption{Deontic}
\end{subtable}
\hfill
\begin{subtable}[h]{0.48\textwidth}
\centering
\resizebox{\linewidth}{!}{%
\begin{tabular}{c c c c c c}
    \toprule
    \textbf{Polarity} & \textbf{Pattern} & \textbf{TA} & \textbf{FA} & \textbf{TC} & \textbf{FC} \\
    \midrule
    \texttt{pos-pos} & $p \to q$ & \fcolorbox{black}{\corC}{100.0} & 10.0 & \fcolorbox{black}{white}{0.0} & \colorbox{\corC}{90.0} \\
    \texttt{pos-neg} & $p \to \lnot q$ & \fcolorbox{black}{\corC}{100.0} & 5.0 & 10.0 & \fcolorbox{black}{\corC}{85.0} \\
    \texttt{neg-pos} & $\lnot p \to q$ & \colorbox{\corC}{90.0} & \fcolorbox{black}{white}{0.0} & \fcolorbox{black}{white}{55.0} & \colorbox{\corC}{100.0} \\
    \texttt{neg-neg} & $\lnot p \to \lnot q$ & \colorbox{\corC}{90.0} & \fcolorbox{black}{white}{0.0} & 55.0 & \fcolorbox{black}{\corC}{100.0} \\
    \bottomrule
\end{tabular}
}
\caption{Descriptive}
\end{subtable}
\caption{qwen3-14b (Few-Shot)}
\label{tab:results-qwen3-14b-kshot}
\end{table}

\begin{table}[H]
\centering
\begin{subtable}[h]{0.48\textwidth}
\centering
\resizebox{\linewidth}{!}{%
\begin{tabular}{c c c c c c}
    \toprule
    \textbf{Polarity} & \textbf{Pattern} & \textbf{TA} & \textbf{FA} & \textbf{TC} & \textbf{FC} \\
    \midrule
    \texttt{pos-pos} & $p \to q$ & \fcolorbox{black}{\corC}{100.0} & 0.0 & \fcolorbox{black}{white}{0.0} & \colorbox{\corC}{100.0} \\
    \texttt{pos-neg} & $p \to \lnot q$ & \fcolorbox{black}{\corC}{100.0} & 0.0 & 0.0 & \fcolorbox{black}{\corC}{100.0} \\
    \texttt{neg-pos} & $\lnot p \to q$ & \colorbox{\corC}{100.0} & \fcolorbox{black}{white}{0.0} & \fcolorbox{black}{white}{0.0} & \colorbox{\corC}{100.0} \\
    \texttt{neg-neg} & $\lnot p \to \lnot q$ & \colorbox{\corC}{95.0} & \fcolorbox{black}{white}{0.0} & 0.0 & \fcolorbox{black}{\corC}{100.0} \\
    \bottomrule
\end{tabular}
}
\caption{Deontic}
\end{subtable}
\hfill
\begin{subtable}[h]{0.48\textwidth}
\centering
\resizebox{\linewidth}{!}{%
\begin{tabular}{c c c c c c}
    \toprule
    \textbf{Polarity} & \textbf{Pattern} & \textbf{TA} & \textbf{FA} & \textbf{TC} & \textbf{FC} \\
    \midrule
    \texttt{pos-pos} & $p \to q$ & \fcolorbox{black}{\corC}{100.0} & 25.0 & \fcolorbox{black}{white}{5.0} & \colorbox{\corC}{85.0} \\
    \texttt{pos-neg} & $p \to \lnot q$ & \fcolorbox{black}{\corC}{100.0} & 0.0 & 10.0 & \fcolorbox{black}{\corC}{85.0} \\
    \texttt{neg-pos} & $\lnot p \to q$ & \colorbox{\corC}{100.0} & \fcolorbox{black}{white}{0.0} & \fcolorbox{black}{white}{30.0} & \colorbox{\corC}{95.0} \\
    \texttt{neg-neg} & $\lnot p \to \lnot q$ & \colorbox{\corC}{100.0} & \fcolorbox{black}{white}{0.0} & 55.0 & \fcolorbox{black}{\corC}{95.0} \\
    \bottomrule
\end{tabular}
}
\caption{Descriptive}
\end{subtable}
\caption{qwen3-32b (Zero-Shot)}
\label{tab:results-qwen3-32b-base}
\end{table}

\begin{table}[H]
\centering
\begin{subtable}[h]{0.48\textwidth}
\centering
\resizebox{\linewidth}{!}{%
\begin{tabular}{c c c c c c}
    \toprule
    \textbf{Polarity} & \textbf{Pattern} & \textbf{TA} & \textbf{FA} & \textbf{TC} & \textbf{FC} \\
    \midrule
    \texttt{pos-pos} & $p \to q$ & \fcolorbox{black}{\corC}{100.0} & 0.0 & \fcolorbox{black}{white}{0.0} & \colorbox{\corC}{100.0} \\
    \texttt{pos-neg} & $p \to \lnot q$ & \fcolorbox{black}{\corC}{100.0} & 0.0 & 0.0 & \fcolorbox{black}{\corC}{100.0} \\
    \texttt{neg-pos} & $\lnot p \to q$ & \colorbox{\corC}{100.0} & \fcolorbox{black}{white}{0.0} & \fcolorbox{black}{white}{0.0} & \colorbox{\corC}{100.0} \\
    \texttt{neg-neg} & $\lnot p \to \lnot q$ & \colorbox{\corC}{100.0} & \fcolorbox{black}{white}{0.0} & 0.0 & \fcolorbox{black}{\corC}{100.0} \\
    \bottomrule
\end{tabular}
}
\caption{Deontic}
\end{subtable}
\hfill
\begin{subtable}[h]{0.48\textwidth}
\centering
\resizebox{\linewidth}{!}{%
\begin{tabular}{c c c c c c}
    \toprule
    \textbf{Polarity} & \textbf{Pattern} & \textbf{TA} & \textbf{FA} & \textbf{TC} & \textbf{FC} \\
    \midrule
    \texttt{pos-pos} & $p \to q$ & \fcolorbox{black}{\corC}{100.0} & 5.0 & \fcolorbox{black}{white}{5.0} & \colorbox{\corC}{95.0} \\
    \texttt{pos-neg} & $p \to \lnot q$ & \fcolorbox{black}{\corC}{100.0} & 0.0 & 20.0 & \fcolorbox{black}{\corC}{100.0} \\
    \texttt{neg-pos} & $\lnot p \to q$ & \colorbox{\corC}{85.0} & \fcolorbox{black}{white}{0.0} & \fcolorbox{black}{white}{30.0} & \colorbox{\corC}{90.0} \\
    \texttt{neg-neg} & $\lnot p \to \lnot q$ & \colorbox{\corC}{100.0} & \fcolorbox{black}{white}{0.0} & 50.0 & \fcolorbox{black}{\corC}{100.0} \\
    \bottomrule
\end{tabular}
}
\caption{Descriptive}
\end{subtable}
\caption{qwen3-32b (Few-Shot)}
\label{tab:results-qwen3-32b-kshot}
\end{table}

\begin{table}[H]
\centering
\begin{subtable}[h]{0.48\textwidth}
\centering
\resizebox{\linewidth}{!}{%
\begin{tabular}{c c c c c c}
    \toprule
    \textbf{Polarity} & \textbf{Pattern} & \textbf{TA} & \textbf{FA} & \textbf{TC} & \textbf{FC} \\
    \midrule
    \texttt{pos-pos} & $p \to q$ & \fcolorbox{black}{\corC}{100.0} & 15.0 & \fcolorbox{black}{white}{20.0} & \colorbox{\corC}{0.0} \\
    \texttt{pos-neg} & $p \to \lnot q$ & \fcolorbox{black}{\corC}{100.0} & 20.0 & 0.0 & \fcolorbox{black}{\corC}{50.0} \\
    \texttt{neg-pos} & $\lnot p \to q$ & \colorbox{\corC}{95.0} & \fcolorbox{black}{white}{5.0} & \fcolorbox{black}{white}{10.0} & \colorbox{\corC}{5.0} \\
    \texttt{neg-neg} & $\lnot p \to \lnot q$ & \colorbox{\corC}{95.0} & \fcolorbox{black}{white}{0.0} & 5.0 & \fcolorbox{black}{\corC}{15.0} \\
    \bottomrule
\end{tabular}
}
\caption{Deontic}
\end{subtable}
\hfill
\begin{subtable}[h]{0.48\textwidth}
\centering
\resizebox{\linewidth}{!}{%
\begin{tabular}{c c c c c c}
    \toprule
    \textbf{Polarity} & \textbf{Pattern} & \textbf{TA} & \textbf{FA} & \textbf{TC} & \textbf{FC} \\
    \midrule
    \texttt{pos-pos} & $p \to q$ & \fcolorbox{black}{\corC}{100.0} & 0.0 & \fcolorbox{black}{white}{55.0} & \colorbox{\corC}{0.0} \\
    \texttt{pos-neg} & $p \to \lnot q$ & \fcolorbox{black}{\corC}{100.0} & 0.0 & 0.0 & \fcolorbox{black}{\corC}{40.0} \\
    \texttt{neg-pos} & $\lnot p \to q$ & \colorbox{\corC}{45.0} & \fcolorbox{black}{white}{85.0} & \fcolorbox{black}{white}{40.0} & \colorbox{\corC}{10.0} \\
    \texttt{neg-neg} & $\lnot p \to \lnot q$ & \colorbox{\corC}{50.0} & \fcolorbox{black}{white}{90.0} & 0.0 & \fcolorbox{black}{\corC}{35.0} \\
    \bottomrule
\end{tabular}
}
\caption{Descriptive}
\end{subtable}
\caption{gemma3-4b (Zero-Shot)}
\label{tab:results-gemma3-4b-base}
\end{table}

\begin{table}[H]
\centering
\begin{subtable}[h]{0.48\textwidth}
\centering
\resizebox{\linewidth}{!}{%
\begin{tabular}{c c c c c c}
    \toprule
    \textbf{Polarity} & \textbf{Pattern} & \textbf{TA} & \textbf{FA} & \textbf{TC} & \textbf{FC} \\
    \midrule
    \texttt{pos-pos} & $p \to q$ & \fcolorbox{black}{\corC}{100.0} & 40.0 & \fcolorbox{black}{white}{0.0} & \colorbox{\corC}{5.0} \\
    \texttt{pos-neg} & $p \to \lnot q$ & \fcolorbox{black}{\corC}{100.0} & 55.0 & 5.0 & \fcolorbox{black}{\corC}{5.0} \\
    \texttt{neg-pos} & $\lnot p \to q$ & \colorbox{\corC}{90.0} & \fcolorbox{black}{white}{35.0} & \fcolorbox{black}{white}{5.0} & \colorbox{\corC}{65.0} \\
    \texttt{neg-neg} & $\lnot p \to \lnot q$ & \colorbox{\corC}{95.0} & \fcolorbox{black}{white}{30.0} & 55.0 & \fcolorbox{black}{\corC}{15.0} \\
    \bottomrule
\end{tabular}
}
\caption{Deontic}
\end{subtable}
\hfill
\begin{subtable}[h]{0.48\textwidth}
\centering
\resizebox{\linewidth}{!}{%
\begin{tabular}{c c c c c c}
    \toprule
    \textbf{Polarity} & \textbf{Pattern} & \textbf{TA} & \textbf{FA} & \textbf{TC} & \textbf{FC} \\
    \midrule
    \texttt{pos-pos} & $p \to q$ & \fcolorbox{black}{\corC}{95.0} & 10.0 & \fcolorbox{black}{white}{60.0} & \colorbox{\corC}{10.0} \\
    \texttt{pos-neg} & $p \to \lnot q$ & \fcolorbox{black}{\corC}{90.0} & 30.0 & 35.0 & \fcolorbox{black}{\corC}{20.0} \\
    \texttt{neg-pos} & $\lnot p \to q$ & \colorbox{\corC}{85.0} & \fcolorbox{black}{white}{70.0} & \fcolorbox{black}{white}{15.0} & \colorbox{\corC}{30.0} \\
    \texttt{neg-neg} & $\lnot p \to \lnot q$ & \colorbox{\corC}{90.0} & \fcolorbox{black}{white}{75.0} & 25.0 & \fcolorbox{black}{\corC}{15.0} \\
    \bottomrule
\end{tabular}
}
\caption{Descriptive}
\end{subtable}
\caption{gemma3-4b (Few-Shot)}
\label{tab:results-gemma3-4b-kshot}
\end{table}

\begin{table}[H]
\centering
\begin{subtable}[h]{0.48\textwidth}
\centering
\resizebox{\linewidth}{!}{%
\begin{tabular}{c c c c c c}
    \toprule
    \textbf{Polarity} & \textbf{Pattern} & \textbf{TA} & \textbf{FA} & \textbf{TC} & \textbf{FC} \\
    \midrule
    \texttt{pos-pos} & $p \to q$ & \fcolorbox{black}{\corC}{100.0} & 30.0 & \fcolorbox{black}{white}{65.0} & \colorbox{\corC}{0.0} \\
    \texttt{pos-neg} & $p \to \lnot q$ & \fcolorbox{black}{\corC}{100.0} & 10.0 & 0.0 & \fcolorbox{black}{\corC}{85.0} \\
    \texttt{neg-pos} & $\lnot p \to q$ & \colorbox{\corC}{90.0} & \fcolorbox{black}{white}{15.0} & \fcolorbox{black}{white}{25.0} & \colorbox{\corC}{20.0} \\
    \texttt{neg-neg} & $\lnot p \to \lnot q$ & \colorbox{\corC}{90.0} & \fcolorbox{black}{white}{5.0} & 5.0 & \fcolorbox{black}{\corC}{55.0} \\
    \bottomrule
\end{tabular}
}
\caption{Deontic}
\end{subtable}
\hfill
\begin{subtable}[h]{0.48\textwidth}
\centering
\resizebox{\linewidth}{!}{%
\begin{tabular}{c c c c c c}
    \toprule
    \textbf{Polarity} & \textbf{Pattern} & \textbf{TA} & \textbf{FA} & \textbf{TC} & \textbf{FC} \\
    \midrule
    \texttt{pos-pos} & $p \to q$ & \fcolorbox{black}{\corC}{95.0} & 15.0 & \fcolorbox{black}{white}{80.0} & \colorbox{\corC}{0.0} \\
    \texttt{pos-neg} & $p \to \lnot q$ & \fcolorbox{black}{\corC}{90.0} & 30.0 & 0.0 & \fcolorbox{black}{\corC}{70.0} \\
    \texttt{neg-pos} & $\lnot p \to q$ & \colorbox{\corC}{75.0} & \fcolorbox{black}{white}{65.0} & \fcolorbox{black}{white}{35.0} & \colorbox{\corC}{10.0} \\
    \texttt{neg-neg} & $\lnot p \to \lnot q$ & \colorbox{\corC}{80.0} & \fcolorbox{black}{white}{80.0} & 0.0 & \fcolorbox{black}{\corC}{30.0} \\
    \bottomrule
\end{tabular}
}
\caption{Descriptive}
\end{subtable}
\caption{gemma3-4b (CoT)}
\label{tab:results-gemma3-4b-cot}
\end{table}

\begin{table}[H]
\centering
\begin{subtable}[h]{0.48\textwidth}
\centering
\resizebox{\linewidth}{!}{%
\begin{tabular}{c c c c c c}
    \toprule
    \textbf{Polarity} & \textbf{Pattern} & \textbf{TA} & \textbf{FA} & \textbf{TC} & \textbf{FC} \\
    \midrule
    \texttt{pos-pos} & $p \to q$ & \fcolorbox{black}{\corC}{75.0} & 75.0 & \fcolorbox{black}{white}{20.0} & \colorbox{\corC}{30.0} \\
    \texttt{pos-neg} & $p \to \lnot q$ & \fcolorbox{black}{\corC}{80.0} & 35.0 & 0.0 & \fcolorbox{black}{\corC}{85.0} \\
    \texttt{neg-pos} & $\lnot p \to q$ & \colorbox{\corC}{95.0} & \fcolorbox{black}{white}{10.0} & \fcolorbox{black}{white}{10.0} & \colorbox{\corC}{85.0} \\
    \texttt{neg-neg} & $\lnot p \to \lnot q$ & \colorbox{\corC}{95.0} & \fcolorbox{black}{white}{10.0} & 5.0 & \fcolorbox{black}{\corC}{85.0} \\
    \bottomrule
\end{tabular}
}
\caption{Deontic}
\end{subtable}
\hfill
\begin{subtable}[h]{0.48\textwidth}
\centering
\resizebox{\linewidth}{!}{%
\begin{tabular}{c c c c c c}
    \toprule
    \textbf{Polarity} & \textbf{Pattern} & \textbf{TA} & \textbf{FA} & \textbf{TC} & \textbf{FC} \\
    \midrule
    \texttt{pos-pos} & $p \to q$ & \fcolorbox{black}{\corC}{95.0} & 35.0 & \fcolorbox{black}{white}{40.0} & \colorbox{\corC}{5.0} \\
    \texttt{pos-neg} & $p \to \lnot q$ & \fcolorbox{black}{\corC}{100.0} & 30.0 & 5.0 & \fcolorbox{black}{\corC}{65.0} \\
    \texttt{neg-pos} & $\lnot p \to q$ & \colorbox{\corC}{90.0} & \fcolorbox{black}{white}{30.0} & \fcolorbox{black}{white}{20.0} & \colorbox{\corC}{35.0} \\
    \texttt{neg-neg} & $\lnot p \to \lnot q$ & \colorbox{\corC}{75.0} & \fcolorbox{black}{white}{45.0} & 35.0 & \fcolorbox{black}{\corC}{45.0} \\
    \bottomrule
\end{tabular}
}
\caption{Descriptive}
\end{subtable}
\caption{gemma3-12b (Zero-Shot)}
\label{tab:results-gemma3-12b-base}
\end{table}

\begin{table}[H]
\centering
\begin{subtable}[h]{0.48\textwidth}
\centering
\resizebox{\linewidth}{!}{%
\begin{tabular}{c c c c c c}
    \toprule
    \textbf{Polarity} & \textbf{Pattern} & \textbf{TA} & \textbf{FA} & \textbf{TC} & \textbf{FC} \\
    \midrule
    \texttt{pos-pos} & $p \to q$ & \fcolorbox{black}{\corC}{95.0} & 20.0 & \fcolorbox{black}{white}{0.0} & \colorbox{\corC}{75.0} \\
    \texttt{pos-neg} & $p \to \lnot q$ & \fcolorbox{black}{\corC}{75.0} & 0.0 & 20.0 & \fcolorbox{black}{\corC}{95.0} \\
    \texttt{neg-pos} & $\lnot p \to q$ & \colorbox{\corC}{90.0} & \fcolorbox{black}{white}{15.0} & \fcolorbox{black}{white}{5.0} & \colorbox{\corC}{90.0} \\
    \texttt{neg-neg} & $\lnot p \to \lnot q$ & \colorbox{\corC}{90.0} & \fcolorbox{black}{white}{5.0} & 15.0 & \fcolorbox{black}{\corC}{90.0} \\
    \bottomrule
\end{tabular}
}
\caption{Deontic}
\end{subtable}
\hfill
\begin{subtable}[h]{0.48\textwidth}
\centering
\resizebox{\linewidth}{!}{%
\begin{tabular}{c c c c c c}
    \toprule
    \textbf{Polarity} & \textbf{Pattern} & \textbf{TA} & \textbf{FA} & \textbf{TC} & \textbf{FC} \\
    \midrule
    \texttt{pos-pos} & $p \to q$ & \fcolorbox{black}{\corC}{75.0} & 30.0 & \fcolorbox{black}{white}{20.0} & \colorbox{\corC}{40.0} \\
    \texttt{pos-neg} & $p \to \lnot q$ & \fcolorbox{black}{\corC}{90.0} & 20.0 & 25.0 & \fcolorbox{black}{\corC}{75.0} \\
    \texttt{neg-pos} & $\lnot p \to q$ & \colorbox{\corC}{75.0} & \fcolorbox{black}{white}{25.0} & \fcolorbox{black}{white}{10.0} & \colorbox{\corC}{90.0} \\
    \texttt{neg-neg} & $\lnot p \to \lnot q$ & \colorbox{\corC}{80.0} & \fcolorbox{black}{white}{30.0} & 80.0 & \fcolorbox{black}{\corC}{65.0} \\
    \bottomrule
\end{tabular}
}
\caption{Descriptive}
\end{subtable}
\caption{gemma3-12b (Few-Shot)}
\label{tab:results-gemma3-12b-kshot}
\end{table}

\begin{table}[H]
\centering
\begin{subtable}[h]{0.48\textwidth}
\centering
\resizebox{\linewidth}{!}{%
\begin{tabular}{c c c c c c}
    \toprule
    \textbf{Polarity} & \textbf{Pattern} & \textbf{TA} & \textbf{FA} & \textbf{TC} & \textbf{FC} \\
    \midrule
    \texttt{pos-pos} & $p \to q$ & \fcolorbox{black}{\corC}{90.0} & 85.0 & \fcolorbox{black}{white}{5.0} & \colorbox{\corC}{10.0} \\
    \texttt{pos-neg} & $p \to \lnot q$ & \fcolorbox{black}{\corC}{90.0} & 45.0 & 5.0 & \fcolorbox{black}{\corC}{55.0} \\
    \texttt{neg-pos} & $\lnot p \to q$ & \colorbox{\corC}{100.0} & \fcolorbox{black}{white}{0.0} & \fcolorbox{black}{white}{5.0} & \colorbox{\corC}{55.0} \\
    \texttt{neg-neg} & $\lnot p \to \lnot q$ & \colorbox{\corC}{100.0} & \fcolorbox{black}{white}{0.0} & 15.0 & \fcolorbox{black}{\corC}{75.0} \\
    \bottomrule
\end{tabular}
}
\caption{Deontic}
\end{subtable}
\hfill
\begin{subtable}[h]{0.48\textwidth}
\centering
\resizebox{\linewidth}{!}{%
\begin{tabular}{c c c c c c}
    \toprule
    \textbf{Polarity} & \textbf{Pattern} & \textbf{TA} & \textbf{FA} & \textbf{TC} & \textbf{FC} \\
    \midrule
    \texttt{pos-pos} & $p \to q$ & \fcolorbox{black}{\corC}{100.0} & 40.0 & \fcolorbox{black}{white}{0.0} & \colorbox{\corC}{5.0} \\
    \texttt{pos-neg} & $p \to \lnot q$ & \fcolorbox{black}{\corC}{100.0} & 55.0 & 10.0 & \fcolorbox{black}{\corC}{15.0} \\
    \texttt{neg-pos} & $\lnot p \to q$ & \colorbox{\corC}{95.0} & \fcolorbox{black}{white}{30.0} & \fcolorbox{black}{white}{10.0} & \colorbox{\corC}{30.0} \\
    \texttt{neg-neg} & $\lnot p \to \lnot q$ & \colorbox{\corC}{95.0} & \fcolorbox{black}{white}{20.0} & 40.0 & \fcolorbox{black}{\corC}{20.0} \\
    \bottomrule
\end{tabular}
}
\caption{Descriptive}
\end{subtable}
\caption{gemma3-12b (CoT)}
\label{tab:results-gemma3-12b-cot}
\end{table}

\begin{table}[H]
\centering
\begin{subtable}[h]{0.48\textwidth}
\centering
\resizebox{\linewidth}{!}{%
\begin{tabular}{c c c c c c}
    \toprule
    \textbf{Polarity} & \textbf{Pattern} & \textbf{TA} & \textbf{FA} & \textbf{TC} & \textbf{FC} \\
    \midrule
    \texttt{pos-pos} & $p \to q$ & \fcolorbox{black}{\corC}{100.0} & 10.0 & \fcolorbox{black}{white}{15.0} & \colorbox{\corC}{75.0} \\
    \texttt{pos-neg} & $p \to \lnot q$ & \fcolorbox{black}{\corC}{100.0} & 0.0 & 0.0 & \fcolorbox{black}{\corC}{100.0} \\
    \texttt{neg-pos} & $\lnot p \to q$ & \colorbox{\corC}{85.0} & \fcolorbox{black}{white}{30.0} & \fcolorbox{black}{white}{10.0} & \colorbox{\corC}{75.0} \\
    \texttt{neg-neg} & $\lnot p \to \lnot q$ & \colorbox{\corC}{85.0} & \fcolorbox{black}{white}{20.0} & 15.0 & \fcolorbox{black}{\corC}{85.0} \\
    \bottomrule
\end{tabular}
}
\caption{Deontic}
\end{subtable}
\hfill
\begin{subtable}[h]{0.48\textwidth}
\centering
\resizebox{\linewidth}{!}{%
\begin{tabular}{c c c c c c}
    \toprule
    \textbf{Polarity} & \textbf{Pattern} & \textbf{TA} & \textbf{FA} & \textbf{TC} & \textbf{FC} \\
    \midrule
    \texttt{pos-pos} & $p \to q$ & \fcolorbox{black}{\corC}{100.0} & 10.0 & \fcolorbox{black}{white}{90.0} & \colorbox{\corC}{0.0} \\
    \texttt{pos-neg} & $p \to \lnot q$ & \fcolorbox{black}{\corC}{100.0} & 5.0 & 0.0 & \fcolorbox{black}{\corC}{95.0} \\
    \texttt{neg-pos} & $\lnot p \to q$ & \colorbox{\corC}{95.0} & \fcolorbox{black}{white}{45.0} & \fcolorbox{black}{white}{50.0} & \colorbox{\corC}{10.0} \\
    \texttt{neg-neg} & $\lnot p \to \lnot q$ & \colorbox{\corC}{85.0} & \fcolorbox{black}{white}{45.0} & 5.0 & \fcolorbox{black}{\corC}{65.0} \\
    \bottomrule
\end{tabular}
}
\caption{Descriptive}
\end{subtable}
\caption{gemma3-27b (Zero-Shot)}
\label{tab:results-gemma3-27b-base}
\end{table}

\begin{table}[H]
\centering
\begin{subtable}[h]{0.48\textwidth}
\centering
\resizebox{\linewidth}{!}{%
\begin{tabular}{c c c c c c}
    \toprule
    \textbf{Polarity} & \textbf{Pattern} & \textbf{TA} & \textbf{FA} & \textbf{TC} & \textbf{FC} \\
    \midrule
    \texttt{pos-pos} & $p \to q$ & \fcolorbox{black}{\corC}{100.0} & 10.0 & \fcolorbox{black}{white}{0.0} & \colorbox{\corC}{90.0} \\
    \texttt{pos-neg} & $p \to \lnot q$ & \fcolorbox{black}{\corC}{100.0} & 0.0 & 0.0 & \fcolorbox{black}{\corC}{100.0} \\
    \texttt{neg-pos} & $\lnot p \to q$ & \colorbox{\corC}{70.0} & \fcolorbox{black}{white}{35.0} & \fcolorbox{black}{white}{30.0} & \colorbox{\corC}{65.0} \\
    \texttt{neg-neg} & $\lnot p \to \lnot q$ & \colorbox{\corC}{75.0} & \fcolorbox{black}{white}{25.0} & 0.0 & \fcolorbox{black}{\corC}{100.0} \\
    \bottomrule
\end{tabular}
}
\caption{Deontic}
\end{subtable}
\hfill
\begin{subtable}[h]{0.48\textwidth}
\centering
\resizebox{\linewidth}{!}{%
\begin{tabular}{c c c c c c}
    \toprule
    \textbf{Polarity} & \textbf{Pattern} & \textbf{TA} & \textbf{FA} & \textbf{TC} & \textbf{FC} \\
    \midrule
    \texttt{pos-pos} & $p \to q$ & \fcolorbox{black}{\corC}{100.0} & 85.0 & \fcolorbox{black}{white}{5.0} & \colorbox{\corC}{10.0} \\
    \texttt{pos-neg} & $p \to \lnot q$ & \fcolorbox{black}{\corC}{100.0} & 10.0 & 0.0 & \fcolorbox{black}{\corC}{90.0} \\
    \texttt{neg-pos} & $\lnot p \to q$ & \colorbox{\corC}{100.0} & \fcolorbox{black}{white}{35.0} & \fcolorbox{black}{white}{25.0} & \colorbox{\corC}{40.0} \\
    \texttt{neg-neg} & $\lnot p \to \lnot q$ & \colorbox{\corC}{70.0} & \fcolorbox{black}{white}{35.0} & 15.0 & \fcolorbox{black}{\corC}{80.0} \\
    \bottomrule
\end{tabular}
}
\caption{Descriptive}
\end{subtable}
\caption{gemma3-27b (Few-Shot)}
\label{tab:results-gemma3-27b-kshot}
\end{table}

\begin{table}[H]
\centering
\begin{subtable}[h]{0.48\textwidth}
\centering
\resizebox{\linewidth}{!}{%
\begin{tabular}{c c c c c c}
    \toprule
    \textbf{Polarity} & \textbf{Pattern} & \textbf{TA} & \textbf{FA} & \textbf{TC} & \textbf{FC} \\
    \midrule
    \texttt{pos-pos} & $p \to q$ & \fcolorbox{black}{\corC}{100.0} & 5.0 & \fcolorbox{black}{white}{65.0} & \colorbox{\corC}{30.0} \\
    \texttt{pos-neg} & $p \to \lnot q$ & \fcolorbox{black}{\corC}{100.0} & 0.0 & 0.0 & \fcolorbox{black}{\corC}{100.0} \\
    \texttt{neg-pos} & $\lnot p \to q$ & \colorbox{\corC}{85.0} & \fcolorbox{black}{white}{30.0} & \fcolorbox{black}{white}{30.0} & \colorbox{\corC}{55.0} \\
    \texttt{neg-neg} & $\lnot p \to \lnot q$ & \colorbox{\corC}{90.0} & \fcolorbox{black}{white}{15.0} & 5.0 & \fcolorbox{black}{\corC}{90.0} \\
    \bottomrule
\end{tabular}
}
\caption{Deontic}
\end{subtable}
\hfill
\begin{subtable}[h]{0.48\textwidth}
\centering
\resizebox{\linewidth}{!}{%
\begin{tabular}{c c c c c c}
    \toprule
    \textbf{Polarity} & \textbf{Pattern} & \textbf{TA} & \textbf{FA} & \textbf{TC} & \textbf{FC} \\
    \midrule
    \texttt{pos-pos} & $p \to q$ & \fcolorbox{black}{\corC}{100.0} & 10.0 & \fcolorbox{black}{white}{90.0} & \colorbox{\corC}{0.0} \\
    \texttt{pos-neg} & $p \to \lnot q$ & \fcolorbox{black}{\corC}{100.0} & 5.0 & 0.0 & \fcolorbox{black}{\corC}{95.0} \\
    \texttt{neg-pos} & $\lnot p \to q$ & \colorbox{\corC}{90.0} & \fcolorbox{black}{white}{55.0} & \fcolorbox{black}{white}{55.0} & \colorbox{\corC}{0.0} \\
    \texttt{neg-neg} & $\lnot p \to \lnot q$ & \colorbox{\corC}{85.0} & \fcolorbox{black}{white}{45.0} & 0.0 & \fcolorbox{black}{\corC}{70.0} \\
    \bottomrule
\end{tabular}
}
\caption{Descriptive}
\end{subtable}
\caption{gemma3-27b (CoT)}
\label{tab:results-gemma3-27b-cot}
\end{table}

\begin{table}[H]
\centering
\begin{subtable}[h]{0.48\textwidth}
\centering
\resizebox{\linewidth}{!}{%
\begin{tabular}{c c c c c c}
    \toprule
    \textbf{Polarity} & \textbf{Pattern} & \textbf{TA} & \textbf{FA} & \textbf{TC} & \textbf{FC} \\
    \midrule
    \texttt{pos-pos} & $p \to q$ & \fcolorbox{black}{\corC}{100.0} & 0.0 & \fcolorbox{black}{white}{85.0} & \colorbox{\corC}{30.0} \\
    \texttt{pos-neg} & $p \to \lnot q$ & \fcolorbox{black}{\corC}{100.0} & 0.0 & 0.0 & \fcolorbox{black}{\corC}{100.0} \\
    \texttt{neg-pos} & $\lnot p \to q$ & \colorbox{\corC}{85.0} & \fcolorbox{black}{white}{25.0} & \fcolorbox{black}{white}{40.0} & \colorbox{\corC}{65.0} \\
    \texttt{neg-neg} & $\lnot p \to \lnot q$ & \colorbox{\corC}{85.0} & \fcolorbox{black}{white}{15.0} & 0.0 & \fcolorbox{black}{\corC}{100.0} \\
    \bottomrule
\end{tabular}
}
\caption{Deontic}
\end{subtable}
\hfill
\begin{subtable}[h]{0.48\textwidth}
\centering
\resizebox{\linewidth}{!}{%
\begin{tabular}{c c c c c c}
    \toprule
    \textbf{Polarity} & \textbf{Pattern} & \textbf{TA} & \textbf{FA} & \textbf{TC} & \textbf{FC} \\
    \midrule
    \texttt{pos-pos} & $p \to q$ & \fcolorbox{black}{\corC}{100.0} & 60.0 & \fcolorbox{black}{white}{70.0} & \colorbox{\corC}{15.0} \\
    \texttt{pos-neg} & $p \to \lnot q$ & \fcolorbox{black}{\corC}{100.0} & 15.0 & 5.0 & \fcolorbox{black}{\corC}{85.0} \\
    \texttt{neg-pos} & $\lnot p \to q$ & \colorbox{\corC}{90.0} & \fcolorbox{black}{white}{15.0} & \fcolorbox{black}{white}{30.0} & \colorbox{\corC}{75.0} \\
    \texttt{neg-neg} & $\lnot p \to \lnot q$ & \colorbox{\corC}{95.0} & \fcolorbox{black}{white}{0.0} & 25.0 & \fcolorbox{black}{\corC}{70.0} \\
    \bottomrule
\end{tabular}
}
\caption{Descriptive}
\end{subtable}
\caption{llama-3.3-70b (Zero-Shot)}
\label{tab:results-llama-3.3-70b-base}
\end{table}

\begin{table}[H]
\centering
\begin{subtable}[h]{0.48\textwidth}
\centering
\resizebox{\linewidth}{!}{%
\begin{tabular}{c c c c c c}
    \toprule
    \textbf{Polarity} & \textbf{Pattern} & \textbf{TA} & \textbf{FA} & \textbf{TC} & \textbf{FC} \\
    \midrule
    \texttt{pos-pos} & $p \to q$ & \fcolorbox{black}{\corC}{100.0} & 0.0 & \fcolorbox{black}{white}{25.0} & \colorbox{\corC}{80.0} \\
    \texttt{pos-neg} & $p \to \lnot q$ & \fcolorbox{black}{\corC}{100.0} & 0.0 & 0.0 & \fcolorbox{black}{\corC}{100.0} \\
    \texttt{neg-pos} & $\lnot p \to q$ & \colorbox{\corC}{100.0} & \fcolorbox{black}{white}{5.0} & \fcolorbox{black}{white}{55.0} & \colorbox{\corC}{40.0} \\
    \texttt{neg-neg} & $\lnot p \to \lnot q$ & \colorbox{\corC}{100.0} & \fcolorbox{black}{white}{0.0} & 0.0 & \fcolorbox{black}{\corC}{100.0} \\
    \bottomrule
\end{tabular}
}
\caption{Deontic}
\end{subtable}
\hfill
\begin{subtable}[h]{0.48\textwidth}
\centering
\resizebox{\linewidth}{!}{%
\begin{tabular}{c c c c c c}
    \toprule
    \textbf{Polarity} & \textbf{Pattern} & \textbf{TA} & \textbf{FA} & \textbf{TC} & \textbf{FC} \\
    \midrule
    \texttt{pos-pos} & $p \to q$ & \fcolorbox{black}{\corC}{100.0} & 40.0 & \fcolorbox{black}{white}{40.0} & \colorbox{\corC}{25.0} \\
    \texttt{pos-neg} & $p \to \lnot q$ & \fcolorbox{black}{\corC}{95.0} & 10.0 & 0.0 & \fcolorbox{black}{\corC}{95.0} \\
    \texttt{neg-pos} & $\lnot p \to q$ & \colorbox{\corC}{100.0} & \fcolorbox{black}{white}{5.0} & \fcolorbox{black}{white}{5.0} & \colorbox{\corC}{90.0} \\
    \texttt{neg-neg} & $\lnot p \to \lnot q$ & \colorbox{\corC}{95.0} & \fcolorbox{black}{white}{10.0} & 45.0 & \fcolorbox{black}{\corC}{55.0} \\
    \bottomrule
\end{tabular}
}
\caption{Descriptive}
\end{subtable}
\caption{llama-3.3-70b (Few-Shot)}
\label{tab:results-llama-3.3-70b-kshot}
\end{table}

\begin{table}[H]
\centering
\begin{subtable}[h]{0.48\textwidth}
\centering
\resizebox{\linewidth}{!}{%
\begin{tabular}{c c c c c c}
    \toprule
    \textbf{Polarity} & \textbf{Pattern} & \textbf{TA} & \textbf{FA} & \textbf{TC} & \textbf{FC} \\
    \midrule
    \texttt{pos-pos} & $p \to q$ & \fcolorbox{black}{\corC}{100.0} & 0.0 & \fcolorbox{black}{white}{20.0} & \colorbox{\corC}{90.0} \\
    \texttt{pos-neg} & $p \to \lnot q$ & \fcolorbox{black}{\corC}{100.0} & 0.0 & 0.0 & \fcolorbox{black}{\corC}{100.0} \\
    \texttt{neg-pos} & $\lnot p \to q$ & \colorbox{\corC}{95.0} & \fcolorbox{black}{white}{15.0} & \fcolorbox{black}{white}{20.0} & \colorbox{\corC}{95.0} \\
    \texttt{neg-neg} & $\lnot p \to \lnot q$ & \colorbox{\corC}{85.0} & \fcolorbox{black}{white}{15.0} & 0.0 & \fcolorbox{black}{\corC}{100.0} \\
    \bottomrule
\end{tabular}
}
\caption{Deontic}
\end{subtable}
\hfill
\begin{subtable}[h]{0.48\textwidth}
\centering
\resizebox{\linewidth}{!}{%
\begin{tabular}{c c c c c c}
    \toprule
    \textbf{Polarity} & \textbf{Pattern} & \textbf{TA} & \textbf{FA} & \textbf{TC} & \textbf{FC} \\
    \midrule
    \texttt{pos-pos} & $p \to q$ & \fcolorbox{black}{\corC}{100.0} & 10.0 & \fcolorbox{black}{white}{45.0} & \colorbox{\corC}{15.0} \\
    \texttt{pos-neg} & $p \to \lnot q$ & \fcolorbox{black}{\corC}{100.0} & 5.0 & 10.0 & \fcolorbox{black}{\corC}{75.0} \\
    \texttt{neg-pos} & $\lnot p \to q$ & \colorbox{\corC}{80.0} & \fcolorbox{black}{white}{15.0} & \fcolorbox{black}{white}{0.0} & \colorbox{\corC}{80.0} \\
    \texttt{neg-neg} & $\lnot p \to \lnot q$ & \colorbox{\corC}{85.0} & \fcolorbox{black}{white}{20.0} & 35.0 & \fcolorbox{black}{\corC}{50.0} \\
    \bottomrule
\end{tabular}
}
\caption{Descriptive}
\end{subtable}
\caption{llama-3.3-70b (CoT)}
\label{tab:results-llama-3.3-70b-cot}
\end{table}

\begin{table}[H]
\centering
\begin{subtable}[h]{0.48\textwidth}
\centering
\resizebox{\linewidth}{!}{%
\begin{tabular}{c c c c c c}
    \toprule
    \textbf{Polarity} & \textbf{Pattern} & \textbf{TA} & \textbf{FA} & \textbf{TC} & \textbf{FC} \\
    \midrule
    \texttt{pos-pos} & $p \to q$ & \fcolorbox{black}{\corC}{80.0} & 0.0 & \fcolorbox{black}{white}{90.0} & \colorbox{\corC}{10.0} \\
    \texttt{pos-neg} & $p \to \lnot q$ & \fcolorbox{black}{\corC}{60.0} & 0.0 & 0.0 & \fcolorbox{black}{\corC}{100.0} \\
    \texttt{neg-pos} & $\lnot p \to q$ & \colorbox{\corC}{75.0} & \fcolorbox{black}{white}{0.0} & \fcolorbox{black}{white}{80.0} & \colorbox{\corC}{40.0} \\
    \texttt{neg-neg} & $\lnot p \to \lnot q$ & \colorbox{\corC}{55.0} & \fcolorbox{black}{white}{0.0} & 5.0 & \fcolorbox{black}{\corC}{100.0} \\
    \bottomrule
\end{tabular}
}
\caption{Deontic}
\end{subtable}
\hfill
\begin{subtable}[h]{0.48\textwidth}
\centering
\resizebox{\linewidth}{!}{%
\begin{tabular}{c c c c c c}
    \toprule
    \textbf{Polarity} & \textbf{Pattern} & \textbf{TA} & \textbf{FA} & \textbf{TC} & \textbf{FC} \\
    \midrule
    \texttt{pos-pos} & $p \to q$ & \fcolorbox{black}{\corC}{90.0} & 0.0 & \fcolorbox{black}{white}{70.0} & \colorbox{\corC}{15.0} \\
    \texttt{pos-neg} & $p \to \lnot q$ & \fcolorbox{black}{\corC}{85.0} & 0.0 & 10.0 & \fcolorbox{black}{\corC}{75.0} \\
    \texttt{neg-pos} & $\lnot p \to q$ & \colorbox{\corC}{40.0} & \fcolorbox{black}{white}{40.0} & \fcolorbox{black}{white}{65.0} & \colorbox{\corC}{50.0} \\
    \texttt{neg-neg} & $\lnot p \to \lnot q$ & \colorbox{\corC}{15.0} & \fcolorbox{black}{white}{60.0} & 30.0 & \fcolorbox{black}{\corC}{85.0} \\
    \bottomrule
\end{tabular}
}
\caption{Descriptive}
\end{subtable}
\caption{olmo-2-0325-32b (Zero-Shot)}
\label{tab:results-olmo-2-0325-32b-base}
\end{table}

\begin{table}[H]
\centering
\begin{subtable}[h]{0.48\textwidth}
\centering
\resizebox{\linewidth}{!}{%
\begin{tabular}{c c c c c c}
    \toprule
    \textbf{Polarity} & \textbf{Pattern} & \textbf{TA} & \textbf{FA} & \textbf{TC} & \textbf{FC} \\
    \midrule
    \texttt{pos-pos} & $p \to q$ & \fcolorbox{black}{\corC}{0.0} & 0.0 & \fcolorbox{black}{white}{100.0} & \colorbox{\corC}{95.0} \\
    \texttt{pos-neg} & $p \to \lnot q$ & \fcolorbox{black}{\corC}{0.0} & 0.0 & 60.0 & \fcolorbox{black}{\corC}{100.0} \\
    \texttt{neg-pos} & $\lnot p \to q$ & \colorbox{\corC}{90.0} & \fcolorbox{black}{white}{0.0} & \fcolorbox{black}{white}{15.0} & \colorbox{\corC}{95.0} \\
    \texttt{neg-neg} & $\lnot p \to \lnot q$ & \colorbox{\corC}{95.0} & \fcolorbox{black}{white}{0.0} & 25.0 & \fcolorbox{black}{\corC}{75.0} \\
    \bottomrule
\end{tabular}
}
\caption{Deontic}
\end{subtable}
\hfill
\begin{subtable}[h]{0.48\textwidth}
\centering
\resizebox{\linewidth}{!}{%
\begin{tabular}{c c c c c c}
    \toprule
    \textbf{Polarity} & \textbf{Pattern} & \textbf{TA} & \textbf{FA} & \textbf{TC} & \textbf{FC} \\
    \midrule
    \texttt{pos-pos} & $p \to q$ & \fcolorbox{black}{\corC}{10.0} & 5.0 & \fcolorbox{black}{white}{85.0} & \colorbox{\corC}{90.0} \\
    \texttt{pos-neg} & $p \to \lnot q$ & \fcolorbox{black}{\corC}{10.0} & 5.0 & 80.0 & \fcolorbox{black}{\corC}{85.0} \\
    \texttt{neg-pos} & $\lnot p \to q$ & \colorbox{\corC}{75.0} & \fcolorbox{black}{white}{0.0} & \fcolorbox{black}{white}{25.0} & \colorbox{\corC}{100.0} \\
    \texttt{neg-neg} & $\lnot p \to \lnot q$ & \colorbox{\corC}{75.0} & \fcolorbox{black}{white}{0.0} & 95.0 & \fcolorbox{black}{\corC}{30.0} \\
    \bottomrule
\end{tabular}
}
\caption{Descriptive}
\end{subtable}
\caption{olmo-2-0325-32b (Few-Shot)}
\label{tab:results-olmo-2-0325-32b-kshot}
\end{table}

\begin{table}[H]
\centering
\begin{subtable}[h]{0.48\textwidth}
\centering
\resizebox{\linewidth}{!}{%
\begin{tabular}{c c c c c c}
    \toprule
    \textbf{Polarity} & \textbf{Pattern} & \textbf{TA} & \textbf{FA} & \textbf{TC} & \textbf{FC} \\
    \midrule
    \texttt{pos-pos} & $p \to q$ & \fcolorbox{black}{\corC}{100.0} & 0.0 & \fcolorbox{black}{white}{80.0} & \colorbox{\corC}{20.0} \\
    \texttt{pos-neg} & $p \to \lnot q$ & \fcolorbox{black}{\corC}{90.0} & 0.0 & 5.0 & \fcolorbox{black}{\corC}{100.0} \\
    \texttt{neg-pos} & $\lnot p \to q$ & \colorbox{\corC}{90.0} & \fcolorbox{black}{white}{5.0} & \fcolorbox{black}{white}{55.0} & \colorbox{\corC}{50.0} \\
    \texttt{neg-neg} & $\lnot p \to \lnot q$ & \colorbox{\corC}{80.0} & \fcolorbox{black}{white}{5.0} & 15.0 & \fcolorbox{black}{\corC}{100.0} \\
    \bottomrule
\end{tabular}
}
\caption{Deontic}
\end{subtable}
\hfill
\begin{subtable}[h]{0.48\textwidth}
\centering
\resizebox{\linewidth}{!}{%
\begin{tabular}{c c c c c c}
    \toprule
    \textbf{Polarity} & \textbf{Pattern} & \textbf{TA} & \textbf{FA} & \textbf{TC} & \textbf{FC} \\
    \midrule
    \texttt{pos-pos} & $p \to q$ & \fcolorbox{black}{\corC}{95.0} & 5.0 & \fcolorbox{black}{white}{85.0} & \colorbox{\corC}{15.0} \\
    \texttt{pos-neg} & $p \to \lnot q$ & \fcolorbox{black}{\corC}{95.0} & 0.0 & 10.0 & \fcolorbox{black}{\corC}{95.0} \\
    \texttt{neg-pos} & $\lnot p \to q$ & \colorbox{\corC}{80.0} & \fcolorbox{black}{white}{30.0} & \fcolorbox{black}{white}{40.0} & \colorbox{\corC}{50.0} \\
    \texttt{neg-neg} & $\lnot p \to \lnot q$ & \colorbox{\corC}{40.0} & \fcolorbox{black}{white}{65.0} & 10.0 & \fcolorbox{black}{\corC}{90.0} \\
    \bottomrule
\end{tabular}
}
\caption{Descriptive}
\end{subtable}
\caption{olmo-2-0325-32b (CoT)}
\label{tab:results-olmo-2-0325-32b-cot}
\end{table}

\end{document}